\pdfoutput=1   
\documentclass{style}

\usepackage{bm}

\usepackage{xspace}
\usepackage{adjustbox}
\usepackage{subcaption}
\usepackage{wrapfig}
\usepackage{titletoc}
\usepackage{multirow}
\usepackage{float}
\usepackage[round]{natbib}          
\definecolor{linkblue}{RGB}{0,39,245}
\definecolor{hspecred}{HTML}{ee0000}
\hypersetup{colorlinks, linkcolor=linkblue, citecolor=linkblue, urlcolor=linkblue}

\newcommand{\sysname}{H-Spec\xspace}
\DeclareRobustCommand{\secref}[1]{\hyperref[#1]{\textcolor{hspecred}{\S\ref*{#1}}}}

\newcommand{\topstrut}{\rule{0pt}{\dimexpr\arraystretch\ht\strutbox+2.80pt\relax}}
\newcommand{\botstrut}{\rule[-\dimexpr\arraystretch\dp\strutbox+1.72pt\relax]{0pt}{0pt}}

\title{\textcolor{hspecred}{H-Spec}: Parallel Speculative Decoding\\Without a Drafter-Side KV Cache}
\renewcommand\Affilfont{\normalfont\fontsize{10}{12}\selectfont\centering}
\makeatletter\renewcommand\AB@affilsepx{, \protect\Affilfont}\makeatother   
\author[1,$\dagger$]{Weifan Jiang}
\author[2,$\dagger$]{Krishna Teja Chitty-Venkata}
\author[3]{Megan Flynn}
\author[3]{Reed Meyerson}
\author[1]{\authorcr Zhenting Qi}   
\author[1]{Tianyu Wu}
\author[3,4]{Eldar Kurtic}
\author[1]{Minlan Yu}
\author[5,$\dagger$]{Alexandre Marques}
\affil[1]{\textbf{Harvard University}}
\affil[2]{\textbf{Capital One}}
\affil[3]{\textbf{Red Hat}}
\affil[4]{\textbf{ISTA}}
\affil[5]{\textbf{NVIDIA}}
\makeatletter
\renewcommand{\maketitle}{\bgroup\setlength{\parindent}{0pt}\setlength{\parskip}{0pt}
  \vspace*{3pt}
  \begin{center}
    {\titlefont \@title\par}%
    \vskip20pt
    {\@author\par}
    {\normalfont\bfseries\fontsize{10}{12}\selectfont Correspondence: weifanjiang@g.harvard.edu\par}
    \vskip30pt
  \end{center}
  \egroup
  {\abscontent}%
  \thispagestyle{firststyle}
}
\fancypagestyle{firststyle}{
  \fancyhead[L]{\normalfont\normalsize Preprint.}
  \fancyhead[C]{}
  \fancyhead[R]{}
  \fancyfoot[L]{}
  \fancyfoot[C]{}
  \fancyfoot[R]{}
}
\makeatother
\begin{document}

\begin{abstract}
Speculative decoding losslessly accelerates large language model inference by having a lightweight draft model predict future tokens for verification by the target model. Recent block diffusion drafters further reduce drafting latency by predicting multiple tokens in parallel. However, existing block drafters project target hidden states at every input position into a separate drafter-side KV cache, incurring per-request memory and KV-write overhead that grow with concurrency; directly reusing target KVs in place removes this cache but fails to sustain draft quality throughout the block. We propose a hybrid target-context injection method that complements direct target KV reuse with target hidden states only at the last input position, requiring no separate drafter-side KV cache. Building on this design, we propose \sysname, a hybrid Mamba-attention parallel drafter that consumes the two target-context sources through complementary modules. Mamba modules are initialized with projected last-token target hidden states, while attention modules reuse target KVs in place. Despite its recurrent formulation, Mamba's parallel scan allows \sysname to preserve block-parallel drafting. Across three target models and diverse tasks, \sysname improves over the best baseline by 5.0--13.3\% in mean accepted length and 5.3--12.6\% in batch-size-1 inter-token latency speedup. Under concurrent serving, \sysname consistently achieves higher throughput while maintaining lower KV cache utilization than baselines across evaluated concurrency levels.
\end{abstract}

\maketitle
{\renewcommand{\thefootnote}{\fnsymbol{footnote}}\footnotetext[2]{Work done while at Red Hat.}}

\section{Introduction}

Large language models (LLMs) generate text \emph{autoregressively}: each output token is conditioned on all preceding tokens. Speculative decoding~\citep{Leviathan2023Fast} has emerged as a primary solution to this sequential bottleneck: a lightweight draft model predicts the next $k$ tokens; the larger target model then verifies the $k$ tokens in a single forward pass. Traditional drafters predict future tokens autoregressively~\citep{Cheng2024Recurrent, Li2024Eagle, Li2024Eagle2, Li2025Eagle3}; recent works instead leverage \emph{block diffusion models} as drafters to predict $k$ tokens in parallel, further improving speedup~\citep{Chen2026Dflash, Huang2026Domino, Cheng2026Dspark, Karimi2026Dblast, Inco2026Dflash2}.

Block drafters incorporate target-model context through \emph{KV injection}, keeping their predictions closely conditioned on the target model but requiring a separate drafter-side KV cache that increases GPU memory utilization. For every input token, target-model hidden states are projected into Key and Value representations, which the drafter consumes as prefix context when predicting the next block. Since each request maintains its own drafter-side KV cache, both memory usage and KV-write overhead grow under concurrent serving. Therefore, we ask: \textbf{\emph{can we design a parallel drafter that preserves strong target-model conditioning without the additional drafter-side KV cache?}}

The most straightforward way to eliminate this extra cache is to let the block drafter reuse the target model's existing KV cache directly. To evaluate the approach, we modify the KV injection mechanism in a representative block drafter, DFlash~\citep{Chen2026Dflash}, to reuse the target KV cache in place.
Compared to vanilla DFlash, this straightforward solution largely preserves draft acceptance at early draft positions in the block, but consistently degrades at later positions across multiple evaluated tasks (Figure~\ref{fig:dflash-kv-reuse}), suggesting weaker target-context conditioning than hidden-state-based injection. Overall, eliminating this drafter-side KV cache requires more than direct target-KV reuse to sustain draft quality throughout the block.

Motivated by the study above, we propose a hybrid context-injection method that achieves strong target conditioning without the drafter-side KV cache. Specifically, we combine two complementary sources of target-model context: in-place target KV reuse and target-model hidden states only at the last input position. Under causal attention, hidden states at the last input position are conditioned on the entire prefix, providing a full input summary with $\mathcal{O}(1)$ storage; the target model's existing KVs provide fine-grained position-wise information at no additional memory cost. Together, these two sources offer complementary target conditioning without additional $\mathcal{O}(N)$ memory overhead.

Based on this new context injection, we propose \textbf{\sysname}, a hybrid Mamba-attention parallel drafter that consumes the two target-context sources through different modules. The Mamba modules~\citep{Gu2024Mamba, Dao2024Transformers} are initialized with projected target-model hidden states at the last input position, as the prefix summary provided in these hidden states aligns naturally with the role of Mamba's recurrent state. Attention modules consume the target-model KV cache in place, leveraging its position-wise context. Despite its recurrent formulation, Mamba evaluates all draft positions through a parallel scan, allowing \sysname to preserve block-parallel drafting.

\begingroup
\setlength{\intextsep}{0pt}
\begin{wrapfigure}{r}{0.40\columnwidth}
\centering
\includegraphics[width=\linewidth]{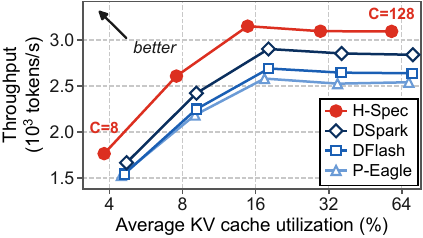}
\captionsetup{skip=1.5pt}
\caption{\label{fig:qwen3-math-tput-kv}\textbf{\sysname achieves the highest throughput and lowest KV utilization} across 8--128 concurrent requests in an example use case.}
\end{wrapfigure}
Evaluations show that \sysname improves both single-request draft quality and end-to-end serving efficiency. Across three target models and diverse tasks, \sysname improves mean accepted length over the best baseline by 5.0--13.3\% and inter-token latency speedup by 5.3--12.6\% at batch size 1, while eliminating the drafter-side KV cache entirely. Under vLLM serving~\citep{Kwon2023Efficient}, \sysname achieves the highest throughput while maintaining the lowest KV cache utilization across concurrency levels. Figure~\ref{fig:qwen3-math-tput-kv} shows an example on a math workload with the Qwen3-8B target model, where \sysname Pareto-dominates other parallel drafters at matched concurrency from 8 to 128 concurrent requests.
\par
\endgroup

To summarize, we identify the drafter-side KV cache as an avoidable memory cost in existing block drafters; we eliminate this bottleneck through a novel hybrid target-context injection mechanism. This design improves both draft quality and serving efficiency. More broadly, our results highlight system efficiency under concurrent serving as an important consideration in drafter design.

\section{Background and motivation}

\noindent\textbf{Speculative decoding speedup.} In each speculative decoding step, a lightweight draft model $\mathcal{M}_D$ proposes $k$ future tokens, which the target model $\mathcal{M}_T$ verifies in one forward pass~\citep{Leviathan2023Fast}. The achieved speedup depends on both the number of accepted draft tokens and the drafter runtime. We discuss the full acceptance procedure in Appendix~\secref{sec:extended-background}. 

Following~\citet{Sadhukhan2025Magicdec}, given $k$ drafted tokens per step, the average per-token latency under speculative decoding is given by: $L_{sd} = (T_D(k) + T_T)/\tau$,
where $T_D(k)$ and $T_T$ represent the time to draft $k$ tokens by $\mathcal{M}_D$, and the time of a single forward pass for $\mathcal{M}_T$, respectively. $1 \leq \tau \leq k+1$ is the mean accepted length (MAL), representing the average number of accepted draft tokens per step plus the target-generated ``bonus'' token. The overall speedup is $L_T / L_{sd}$, where $L_T$ is the per-token latency of $\mathcal{M}_{T}$ without speculative decoding.

\noindent\textbf{Block diffusion for parallel drafting.} Traditional drafters output the next $k$ predictions autoregressively~\citep{Cheng2024Recurrent, Li2024Eagle, Li2024Eagle2, Li2025Eagle3}.
This approach bounds generation speed by the sequential draft cost.
Recent block diffusion drafters break this bottleneck by generating all $k$ draft tokens in a single forward pass. DFlash~\citep{Chen2026Dflash}, a representative block diffusion drafter, uses a stack of Transformer layers with attention and MLP modules. Domino~\citep{Huang2026Domino} and DSpark~\citep{Cheng2026Dspark} further append a lightweight sequential head that refines each draft position based on its preceding information, combining parallel drafting with causal correction.

\begin{figure}[t]
\centering
\begin{minipage}[t]{0.49\textwidth}
\vspace{0pt}
\centering
\includegraphics[width=\linewidth]{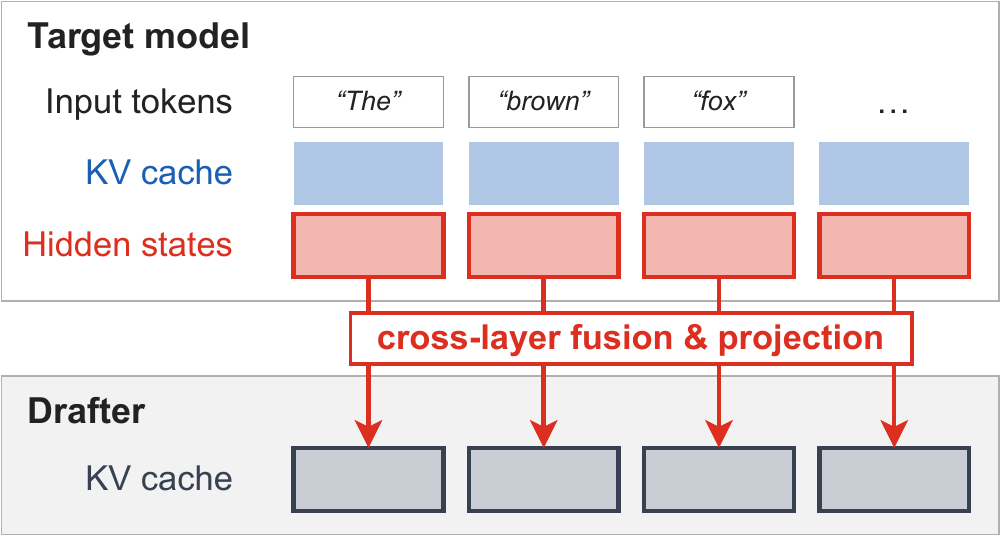}
\captionsetup{skip=5pt}
\captionof{figure}{\label{fig:kv-injection}\textbf{KV injection in existing block diffusion drafters:} Target hidden states at each input position are projected into a drafter-side KV entry, separate from existing target KVs.}
\end{minipage}
\hfill
\begin{minipage}[t]{0.49\textwidth}
\vspace{0pt}
\centering
\captionsetup{skip=1.5pt}
\captionof{table}{\label{tab:dflash-kv-overhead}\textbf{Drafter-side KV introduces additional GPU memory overhead.} Per-token KV cache size across target models and their corresponding official DFlash checkpoints. KV caches are stored in bf16.}
\setlength{\tabcolsep}{3pt}\renewcommand{\arraystretch}{1.15}\fontsize{8}{9.6}\selectfont
\begin{tabular}{@{}l ccc@{}}
\toprule
\textbf{Target model} & \shortstack[c]{\textbf{Target KV}\strut\\\textbf{/token}\strut} & \shortstack[c]{\textbf{Draft KV}\strut\\\textbf{/token}\strut} & \shortstack[c]{\textbf{KV/token}\strut\\\textbf{increase}\strut} \\
\arrayrulecolor{black!38}\midrule\arrayrulecolor{black}
Qwen3-4B             & 144 KiB & 20 KiB & 1.139$\times$ \\
Qwen3-8B             & 144 KiB & 20 KiB & 1.139$\times$ \\
Llama3.1-8B-Instruct & 128 KiB & 20 KiB & 1.156$\times$ \\
Qwen3.5-27B          &  64 KiB & 24 KiB & 1.375$\times$ \\
Qwen3.5-397B-A17B    &  30 KiB & 24 KiB & 1.800$\times$ \\
\bottomrule
\end{tabular}
\end{minipage}
\end{figure}

\noindent\textbf{KV injection in block drafting.} LLM hidden representations capture long-range dependencies and encode information useful for predicting future tokens~\citep{Samragh2025Your}.
Existing block drafters therefore employ a KV injection mechanism that leverages target hidden states (Figure~\ref{fig:kv-injection}). At every input position, hidden states from selected target-model layers are fused and projected into KV representations. The drafter attends to these KVs as prefix context when predicting the next block, keeping draft predictions closely conditioned on the target model and improving acceptance.

\begingroup
\setlength{\intextsep}{0pt}
\begin{wrapfigure}[12]{r}{0.375\columnwidth}
\includegraphics[width=\linewidth]{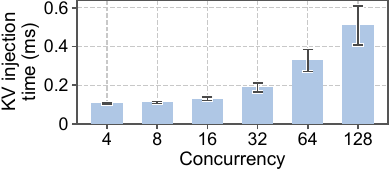}
\captionsetup{skip=3pt}
\caption{\label{fig:kv-overhead-conc}\textbf{Per-step KV injection overhead grows with concurrency}, on the Qwen3-8B target and averaged over five waves; error bars are standard deviations (st.d.).}
\end{wrapfigure}
\noindent\textbf{Limitation: drafter-side KV overhead.} While KV injection provides strong target-context conditioning, it requires an additional drafter-side KV cache in GPU memory for the lifetime of each request. Table~\ref{tab:dflash-kv-overhead} quantifies this cost using the official DFlash checkpoints released by the authors, where the drafter increases per-token KV memory to 1.1--1.8$\times$ compared with serving the target model alone. Furthermore, the memory usage and KV injection overhead increase as the number of in-flight requests grows. As shown in Figure~\ref{fig:kv-overhead-conc}, on Qwen3-8B, per-iteration drafter KV injection time increases by 4.7$\times$ from 4 to 128 concurrent requests.
\par
\endgroup

\section{Method: \sysname}

\subsection{\label{sec:strawman}A straightforward solution: DFlash with direct target-KV reuse}

The most straightforward way to eliminate the drafter-side KV cache is to let the drafter attend directly to the target model's existing KV cache. We evaluate this approach by comparing DFlash with vanilla KV injection and direct target-KV reuse, both trained under a matched recipe.

Vanilla DFlash extracts hidden states from five evenly spaced target-model layers, then fuses and projects them into drafter-side KV representations for the five-layer block diffusion drafter. Our modified version retains the same five-layer structure, but each drafter layer directly reuses the KV cache from one of the five selected target layers. We do not fuse KVs across target layers, as doing so would require materializing fused KVs and reintroduce the drafter-side extra memory.

\begin{figure*}[h]
\includegraphics[width=\columnwidth]{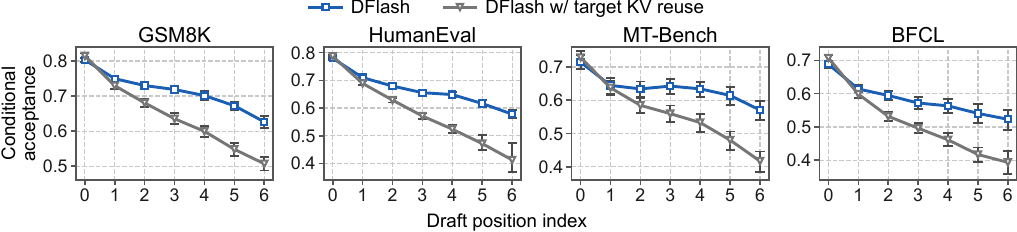}
\captionsetup{skip=3pt}
\caption{\label{fig:dflash-kv-reuse}\textbf{Reusing target KV alone reduces draft quality at later positions.} Per-position conditional acceptance rate for DFlash and the target-KV-reuse variant on Qwen3-8B. Error bars are 95\% confidence intervals (CI) over 10{,}000 request-level bootstrap resamples; other CIs in the paper are computed similarly.}
\end{figure*}

As shown in Figure~\ref{fig:dflash-kv-reuse}, direct target-KV reuse largely retains DFlash's drafting capability at early draft positions (i.e., indices 0 and 1), with differences of $-$2.07 to +1.91 pp. in conditional acceptance rates on math, coding, chat, and agentic tasks. However, the gap widens at later positions, reaching up to 6.3, 12.5, and 16.5 pp. at indices 2, 4, and 6, across the four tasks. Overall, direct target-KV reuse alone does not sustain the draft quality achieved by conventional hidden-state-based KV injection, motivating a richer context injection.

\subsection{\label{sec:hybrid-injection}Hybrid target-context injection}

An important difference between hidden states and KV representations is the contextual information they provide for drafting. At each input position, hidden states integrate information from preceding tokens and encode useful information for predicting future tokens~\citep{Samragh2025Your}. In contrast, target KVs provide position-wise information for retrieval by future queries, but lack an explicit contextual summary over the entire input.

Based on this observation, we propose a novel hybrid target-context injection method that complements direct KV reuse with target-model hidden states only at the last input position. Under causal attention, hidden states at the last position are conditioned on all preceding input tokens, providing the drafter with a compact summary of the full input sequence. Meanwhile, target KVs provide fine-grained position-wise context that complements this compact summary. Importantly, processing and storing hidden states at only the last input position incurs $\mathcal{O}(1)$ overhead relative to input length, eliminating the $\mathcal{O}(N)$ drafter-side KV cache. The remaining challenge is to design a parallel drafter architecture that can make effective use of both context sources.

\subsection{\label{sec:sysname-overview}\sysname architecture}

\begin{figure*}[t]
\centering
\includegraphics[width=\columnwidth]{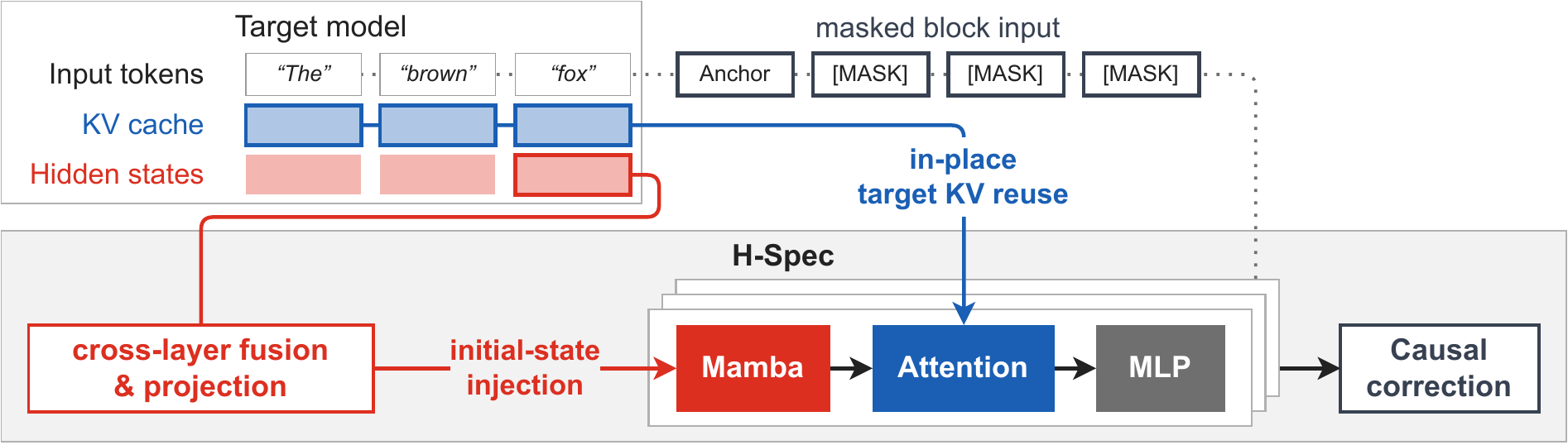}
\captionsetup{aboveskip=5pt,belowskip=-0.5\baselineskip}
\caption{\label{fig:hspec-overview}\textbf{\sysname overview.} \sysname initializes Mamba modules from projected last-token target hidden states and reuses target KVs in place through attention. A lightweight causal correction follows the hybrid backbone to improve inter-token dependencies.}
\end{figure*}

We propose \sysname, a block-parallel drafter built around the hybrid target-context injection method above. Each context source is consumed by a module that aligns with the contextual information it provides. Specifically, the Mamba module is initialized from projected target hidden states at the last input position, whose full-input representation naturally aligns with the role of Mamba's recurrent state. The attention module consumes target KVs in place. Figure~\ref{fig:hspec-overview} illustrates the overall design.

\subsubsection{\label{sec:mamba}Mamba with state initialization from last-token hidden states}

Mamba is a recurrent selective state-space architecture that compresses preceding context into a fixed-size state, which is updated at each position and carried forward to condition subsequent tokens~\citep{Gu2024Mamba}. Given input $x_i$ and preceding state $S_{i-1}$, the state is updated as $S_i = \bar{A}_i S_{i-1} + \bar{B}_i x_i$, where $\bar{A}_i$ and $\bar{B}_i$ are input-dependent. Mamba-2 keeps the underlying selective recurrent formulation but uses a more structured per-head state transition under its State Space Duality (SSD)~\citep{Dao2024Transformers}. \sysname thus uses Mamba-2 for its improved runtime efficiency.

\noindent\textbf{Initial state injection.} As discussed, target hidden states at the last input position are conditioned on the full preceding context, making them naturally aligned with the role of Mamba's recurrent state. We therefore use these hidden states to initialize Mamba: the last-token hidden states are fused across selected target-model layers and projected to serve as Mamba's initial recurrent state, conditioning subsequent draft generation on the target model's full input context.

\noindent\textbf{Cross-layer hidden state fusion.}
Let $h_n^{(m)}$ and $\mathcal{M}$ denote the target-model hidden state at the last input position $n$ from layer $m$, and the set of selected target layers for context injection, respectively. %
\sysname obtains the initial Mamba-2 state $S_{\mathrm{init}}$ via:
\[
F_n(\mathcal{M}) =
\operatorname{Proj}_{r}\!\left(
\operatorname{Concat}\!\left(
\{h_n^{(m)}\}_{m \in \mathcal{M}}
\right)
\right),
\qquad
S_{\mathrm{init}} =
\operatorname{Proj}_{d_S}\!\left(F_n(\mathcal{M})\right)
\]
where $r$ is the intermediate rank used for cross-layer fusion and $d_S$ denotes the flattened dimensionality of Mamba-2's recurrent state. \sysname shares the same $\operatorname{Proj}_{r}$ and $\operatorname{Proj}_{d_S}$ weights across all drafter layers, so the initial state is only computed once per drafter forward pass.

\noindent\textbf{Parallel-drafting compatibility.} Despite its recurrent nature, Mamba can evaluate sequence positions in parallel due to its \emph{linear} state recurrence. In \sysname, the Mamba kernel starts from the injected initial state and computes the state evolution across all draft positions in one pass, avoiding sequential execution. This allows the hybrid Mamba-attention backbone to retain parallel drafting.

\noindent\textbf{Input-dependent selectivity.} Following existing block diffusion drafters, we do not add positional embeddings to the masked block input. Although the first Mamba module receives the same mask-token embedding at each draft position, its recurrent scan produces position-dependent hidden states. Attention modules additionally incorporate positional information through RoPE (\secref{sec:attention}). As a result, Mamba modules after the first receive hidden states that vary across draft positions, preserving input-dependent selectivity across the block.

\subsubsection{\label{sec:attention}Attention modules with in-place target-KV reuse}

In each \sysname layer, the attention module follows the Mamba module, providing fine-grained, position-level context complementary to Mamba's compact recurrent formulation.

\noindent\textbf{KV dimension.} \sysname's attention modules match the target model's KV dimensions, ensuring that target KVs can be reused as is. We also match \sysname's number of attention and KV heads to the target, although in practice reusing only a subset of target KV heads may suffice and could further reduce attention overhead; we leave this design space for future work.

\noindent\textbf{Layer-wise KV mapping.} Each \sysname layer's attention module consumes the KV from one target model layer. This allows \sysname to incorporate target-model information from multiple depths while preserving in-place KV reuse. Cross-layer KV fusion would instead require storing an additional fused KV in memory, reintroducing the drafter KV cache.

\noindent\textbf{Attention mask.} We use causal attention over the masked block, with reused target KVs serving as prefix context to keep the draft closely conditioned on the target model. We use sliding-window attention (SWA) with a 2{,}048-token window for long-context stability~\citep{Eldenk2026Attentiondrift}.

\noindent\textbf{Positional alignment for masked block inputs.} Because target KVs are stored post-RoPE in memory, masked-block Q/K vectors must use the target model's RoPE $\theta$ with position indices that continue from the target prefix, ensuring consistent positional encoding across the prefix and draft block.

\subsubsection{Causal correction}

Block diffusion models process all drafting positions in parallel but can weaken causal dependencies between adjacent positions, reducing coherence in the drafted sequence. Recent works therefore augment the parallel backbone with auxiliary components that strengthen inter-position dependencies. \sysname specifically adopts DSpark's Markov head~\citep{Cheng2026Dspark} after the hybrid backbone, introducing a first-order Markov dependency by adjusting the predicted logits at each draft position based on the prediction at the immediately preceding position.

Causal correction is part of \sysname's architecture but not a contribution of this work, as the hybrid backbone is independent of, and can be paired with, any specific mechanism. We leave evaluating other causal-correction mechanisms with the hybrid backbone as important future work.

\noindent\textbf{Architecture configuration.} To ensure a fair comparison, we match \sysname's trainable and total parameter counts to the recommended 5-layer DFlash/DSpark settings~\citep{Chen2026Dflash, Cheng2026Dspark}. We also match the dimensions of components present in both \sysname and the baselines, including attention, MLP, and the DSpark Markov head. Because each \sysname layer additionally contains a Mamba module, we use a shallower four-layer drafter: three full Mamba-attention-MLP layers and one partial Mamba-MLP layer. Retaining Mamba rather than attention in the partial layer keeps \sysname closest to the baseline parameter budget. We detail the remaining \sysname configurations and dimension settings in Appendix~\secref{sec:sys-config}.

\section{\label{sec:evaluation}Evaluation}

\noindent\textbf{Models and benchmarks.} We conduct experiments on Llama3.1-8B-IT~\citep{Grattafiori2024Llama3} and Qwen3 4B and 8B~\citep{Yang2025Qwen} target models, covering both general-purpose and reasoning LLM families. We evaluate 8 tasks covering diverse domains: math, QA, chat, RAG, summarization, translation, code, and tool calling. The first 6 tasks are adopted from Spec-Bench~\citep{Xia2024Unlocking}, while we add HumanEval~\citep{Chen2021Evaluating} and BFCL~\citep{Patil2025Berkeley} to broaden coding and tool-calling coverage.

\noindent\textbf{Baselines and training configurations.} We compare \sysname with parallel drafters: P-Eagle~\citep{Hui2026Peagle}, DFlash~\citep{Chen2026Dflash}, and DSpark~\citep{Cheng2026Dspark}. Since the official baseline checkpoints use heterogeneous training recipes, we reproduce all baselines and \sysname using Speculators~\citep{Redhat2025Speculators}, the speculative-decoding training framework for the vLLM ecosystem, under a unified training recipe for controlled comparison. We train all drafters for 5 epochs on 100k samples drawn from Magpie~\citep{Xu2025Magpie} and Ultrachat~\citep{Ding2023Ultrachat}, with matched optimizer steps. For Qwen3 targets, we use reasoning-enabled responses generated by the target models for better drafter alignment. Appendix~\secref{sec:exp-config} details remaining baseline and training settings.

\noindent\textbf{Metrics.} We evaluate performance at both the single-request level and under serving load. For single-request, we report mean accepted length (MAL, denoted $\tau$) and inter-token latency (ITL) speedup over no-speculative-decoding at batch size 1. Under serving load, we report throughput (output tokens per second) and KV cache utilization (\%) across varying concurrency levels.

\noindent\textbf{Additional settings.} Unless otherwise specified, we use the default sampling parameters per target model (temp=0.6/topP=0.9 for Llama3.1; temp=0.6/topP=0.95/topK=20 for Qwen3), and 7 drafted tokens per step. We report sensitivity on these settings in \secref{sec:sensitivity}. We measure generation speed and KV cache utilization on one NVIDIA A100 80G GPU.

\subsection{\label{sec:single-request-eval}Single-request accepted length and speedup}

\begin{table*}[h]
\centering
\captionsetup{skip=0pt}
\caption{\textbf{\sysname improves accepted draft length and generation speed under batch size 1.} MAL ($\tau$) and speedup (\textit{Spd.}) at batch size 1 over no-speculative-decoding. Max 4{,}096 tokens per response. \sysname results are tinted; best values are bolded. Statistical uncertainty reported in~\secref{sec:single-request-stat-uncert}.}
\label{tab:rhbench-mal-spd}
\setlength{\tabcolsep}{2.18pt}
\renewcommand{\arraystretch}{1.10}
\fontsize{8}{9.6}\selectfont
\providecommand{\spx}{\kern0.05em{\fontsize{4.6pt}{5pt}\selectfont\texttimes}}
\setlength{\arrayrulewidth}{0.4pt}\setlength{\doublerulesep}{1.8pt}\arrayrulecolor{black!42}
\begin{adjustbox}{width=\linewidth}
\begin{tabular}{l cc|cc|cc|cc|cc|cc|cc|cc|cc}
\arrayrulecolor{black}\specialrule{\heavyrulewidth}{0pt}{0pt}\arrayrulecolor{black!42}
\rule{0pt}{2.6ex}\rule[-0.9ex]{0pt}{0pt} & \multicolumn{2}{c|}{\textbf{Math}} & \multicolumn{2}{c|}{\textbf{QA}} & \multicolumn{2}{c|}{\textbf{Chat}} & \multicolumn{2}{c|}{\textbf{RAG}} & \multicolumn{2}{c|}{\textbf{Summ.}} & \multicolumn{2}{c|}{\textbf{Transl.}} & \multicolumn{2}{c|}{\textbf{Code}} & \multicolumn{2}{c|}{\textbf{Tool}} & \multicolumn{2}{c}{\textbf{Avg.}} \\
\arrayrulecolor{black!60}\specialrule{0.7pt}{0pt}{0pt}\arrayrulecolor{black!42}
\rule{0pt}{2.7ex}{\scshape\color{black!62}L-8B-IT}\rule[-0.8ex]{0pt}{0pt} & {\scriptsize\color{black!62}$\tau$} & {\scriptsize\color{black!62}\textit{Spd.}} & {\scriptsize\color{black!62}$\tau$} & {\scriptsize\color{black!62}\textit{Spd.}} & {\scriptsize\color{black!62}$\tau$} & {\scriptsize\color{black!62}\textit{Spd.}} & {\scriptsize\color{black!62}$\tau$} & {\scriptsize\color{black!62}\textit{Spd.}} & {\scriptsize\color{black!62}$\tau$} & {\scriptsize\color{black!62}\textit{Spd.}} & {\scriptsize\color{black!62}$\tau$} & {\scriptsize\color{black!62}\textit{Spd.}} & {\scriptsize\color{black!62}$\tau$} & {\scriptsize\color{black!62}\textit{Spd.}} & {\scriptsize\color{black!62}$\tau$} & {\scriptsize\color{black!62}\textit{Spd.}} & {\scriptsize\color{black!62}$\tau$} & {\scriptsize\color{black!62}\textit{Spd.}} \\
\arrayrulecolor{black!30}\specialrule{0.3pt}{0pt}{0pt}\arrayrulecolor{black!42}
\rule{0pt}{2.6ex}P-Eagle & 3.08 & \textit{2.82}\spx & \textbf{2.73} & {\bfseries\itshape 2.34}\spx & 2.59 & \textit{2.14}\spx & 2.62 & \textit{2.02}\spx & 2.44 & \textit{1.89}\spx & 1.66 & \textit{1.35}\spx & 3.62 & \textit{2.59}\spx & 2.78 & \textit{2.12}\spx & 2.69 & \textit{2.16}\spx \\
DFlash & 2.99 & \textit{2.57}\spx & 2.19 & \textit{1.65}\spx & 2.69 & \textit{2.30}\spx & 2.29 & \textit{1.83}\spx & 2.11 & \textit{1.67}\spx & 1.44 & \textit{1.20}\spx & 3.46 & \textit{2.88}\spx & 2.63 & \textit{2.09}\spx & 2.47 & \textit{2.03}\spx \\
DSpark & 3.27 & \textit{2.77}\spx & 2.63 & \textit{2.32}\spx & \textbf{3.03} & \textit{2.49}\spx & 2.48 & \textit{1.95}\spx & 2.27 & \textit{1.79}\spx & 1.46 & \textit{1.20}\spx & 3.82 & \textit{3.09}\spx & 2.77 & \textit{2.23}\spx & 2.72 & \textit{2.23}\spx \\
\rowcolor{gray!12}\sysname & \textbf{4.24} & {\bfseries\itshape 3.23}\spx & 2.40 & \textit{2.21}\spx & 2.99 & {\bfseries\itshape 2.64}\spx & \textbf{2.96} & {\bfseries\itshape 2.31}\spx & \textbf{2.77} & {\bfseries\itshape 2.21}\spx & \textbf{1.80} & {\bfseries\itshape 1.43}\spx & \textbf{4.29} & {\bfseries\itshape 3.53}\spx & \textbf{3.19} & {\bfseries\itshape 2.54}\spx & \textbf{3.08} & {\bfseries\itshape 2.51}\spx\rule[-0.8ex]{0pt}{0pt} \\
\arrayrulecolor{black!60}\specialrule{0.7pt}{0pt}{0pt}\arrayrulecolor{black!42}
\rule{0pt}{2.7ex}{\scshape\color{black!62}Q-4B}\rule[-0.8ex]{0pt}{0pt} & {\scriptsize\color{black!62}$\tau$} & {\scriptsize\color{black!62}\textit{Spd.}} & {\scriptsize\color{black!62}$\tau$} & {\scriptsize\color{black!62}\textit{Spd.}} & {\scriptsize\color{black!62}$\tau$} & {\scriptsize\color{black!62}\textit{Spd.}} & {\scriptsize\color{black!62}$\tau$} & {\scriptsize\color{black!62}\textit{Spd.}} & {\scriptsize\color{black!62}$\tau$} & {\scriptsize\color{black!62}\textit{Spd.}} & {\scriptsize\color{black!62}$\tau$} & {\scriptsize\color{black!62}\textit{Spd.}} & {\scriptsize\color{black!62}$\tau$} & {\scriptsize\color{black!62}\textit{Spd.}} & {\scriptsize\color{black!62}$\tau$} & {\scriptsize\color{black!62}\textit{Spd.}} & {\scriptsize\color{black!62}$\tau$} & {\scriptsize\color{black!62}\textit{Spd.}} \\
\arrayrulecolor{black!30}\specialrule{0.3pt}{0pt}{0pt}\arrayrulecolor{black!42}
\rule{0pt}{2.6ex}P-Eagle & 3.76 & \textit{2.77}\spx & 2.84 & \textit{2.15}\spx & 3.04 & \textit{2.18}\spx & 2.92 & \textit{2.08}\spx & 2.39 & \textit{1.74}\spx & 2.69 & \textit{2.10}\spx & 3.41 & \textit{2.37}\spx & 2.87 & \textit{2.08}\spx & 2.99 & \textit{2.18}\spx \\
DFlash & 3.74 & \textit{2.93}\spx & 2.67 & \textit{2.22}\spx & 2.91 & \textit{2.31}\spx & 2.68 & \textit{2.12}\spx & 2.28 & \textit{1.80}\spx & 2.43 & \textit{2.03}\spx & 3.39 & \textit{2.52}\spx & 2.80 & \textit{2.20}\spx & 2.86 & \textit{2.26}\spx \\
DSpark & 4.02 & \textit{3.04}\spx & 2.90 & \textit{2.31}\spx & 3.16 & \textit{2.43}\spx & 2.92 & \textit{2.22}\spx & 2.45 & \textit{1.87}\spx & 2.58 & \textit{2.11}\spx & 3.69 & \textit{2.70}\spx & 2.99 & \textit{2.30}\spx & 3.09 & \textit{2.37}\spx \\
\rowcolor{gray!12}\sysname & \textbf{4.17} & {\bfseries\itshape 3.19}\spx & \textbf{3.04} & {\bfseries\itshape 2.43}\spx & \textbf{3.32} & {\bfseries\itshape 2.58}\spx & \textbf{3.12} & {\bfseries\itshape 2.32}\spx & \textbf{2.62} & {\bfseries\itshape 2.01}\spx & \textbf{2.74} & {\bfseries\itshape 2.23}\spx & \textbf{3.83} & {\bfseries\itshape 2.82}\spx & \textbf{3.09} & {\bfseries\itshape 2.42}\spx & \textbf{3.24} & {\bfseries\itshape 2.50}\spx\rule[-0.8ex]{0pt}{0pt} \\
\arrayrulecolor{black!60}\specialrule{0.7pt}{0pt}{0pt}\arrayrulecolor{black!42}
\rule{0pt}{2.7ex}{\scshape\color{black!62}Q-8B}\rule[-0.8ex]{0pt}{0pt} & {\scriptsize\color{black!62}$\tau$} & {\scriptsize\color{black!62}\textit{Spd.}} & {\scriptsize\color{black!62}$\tau$} & {\scriptsize\color{black!62}\textit{Spd.}} & {\scriptsize\color{black!62}$\tau$} & {\scriptsize\color{black!62}\textit{Spd.}} & {\scriptsize\color{black!62}$\tau$} & {\scriptsize\color{black!62}\textit{Spd.}} & {\scriptsize\color{black!62}$\tau$} & {\scriptsize\color{black!62}\textit{Spd.}} & {\scriptsize\color{black!62}$\tau$} & {\scriptsize\color{black!62}\textit{Spd.}} & {\scriptsize\color{black!62}$\tau$} & {\scriptsize\color{black!62}\textit{Spd.}} & {\scriptsize\color{black!62}$\tau$} & {\scriptsize\color{black!62}\textit{Spd.}} & {\scriptsize\color{black!62}$\tau$} & {\scriptsize\color{black!62}\textit{Spd.}} \\
\arrayrulecolor{black!30}\specialrule{0.3pt}{0pt}{0pt}\arrayrulecolor{black!42}
\rule{0pt}{2.6ex}P-Eagle & 3.74 & \textit{2.91}\spx & 2.65 & \textit{2.19}\spx & 2.98 & \textit{2.33}\spx & 2.83 & \textit{2.16}\spx & 2.36 & \textit{1.88}\spx & 2.63 & \textit{2.14}\spx & 3.36 & \textit{2.56}\spx & 2.83 & \textit{2.22}\spx & 2.92 & \textit{2.30}\spx \\
DFlash & 3.68 & \textit{2.97}\spx & 2.58 & \textit{2.17}\spx & 2.85 & \textit{2.35}\spx & 2.58 & \textit{2.11}\spx & 2.22 & \textit{1.82}\spx & 2.31 & \textit{1.99}\spx & 3.29 & \textit{2.60}\spx & 2.68 & \textit{2.21}\spx & 2.77 & \textit{2.28}\spx \\
DSpark & 3.91 & \textit{3.17}\spx & 2.69 & \textit{2.27}\spx & 3.11 & \textit{2.52}\spx & 2.74 & \textit{2.21}\spx & 2.34 & \textit{1.89}\spx & 2.42 & \textit{2.03}\spx & 3.57 & \textit{2.81}\spx & 2.84 & \textit{2.32}\spx & 2.95 & \textit{2.40}\spx \\
\rowcolor{gray!12}\sysname & \textbf{4.13} & {\bfseries\itshape 3.36}\spx & \textbf{2.97} & {\bfseries\itshape 2.44}\spx & \textbf{3.32} & {\bfseries\itshape 2.69}\spx & \textbf{3.03} & {\bfseries\itshape 2.46}\spx & \textbf{2.63} & {\bfseries\itshape 2.12}\spx & \textbf{2.74} & {\bfseries\itshape 2.31}\spx & \textbf{3.82} & {\bfseries\itshape 3.00}\spx & \textbf{3.06} & {\bfseries\itshape 2.50}\spx & \textbf{3.21} & {\bfseries\itshape 2.61}\spx\rule[-0.8ex]{0pt}{0pt} \\
\arrayrulecolor{black}\specialrule{\heavyrulewidth}{0pt}{0pt}\arrayrulecolor{black!42}
\end{tabular}
\end{adjustbox}\arrayrulecolor{black}
\end{table*}

As shown in Table~\ref{tab:rhbench-mal-spd}, \sysname improves the average unweighted $\tau$ across eight tasks by 13.3\%, 5.0\%, and 8.7\%, and Spd. by 12.6\%, 5.3\%, and 8.7\% over the best baseline for Llama, Qwen3 4B, and 8B targets, respectively. Paired bootstrap 95\% CIs for all six comparisons sit above zero, indicating that \sysname's gains are statistically significant (Appendix~\secref{sec:single-request-stat-uncert}). Notably, these gains are achieved while \sysname requires \emph{no} drafter-side KV cache by design, whereas all baselines require it.

\noindent\textbf{Extended evaluation tasks.} Appendix~\secref{sec:deepspec-eval} evaluates the 9-task suite released by the DSpark authors. Across three target models, \sysname improves average $\tau$ by 3.8--12.5\%, and Spd. by 4.3--13.1\% over the best baseline. Importantly, \sysname leads in both $\tau$ and Spd. on every task for all target models with statistical significance, demonstrating consistent cross-domain gains.

\subsection{\label{sec:serving-eval}Serving performance using vLLM}

We evaluate \sysname's serving performance using vLLM~\citep{Kwon2023Efficient}. Following DSpark's evaluation, we use one workload for each of the math, coding, and chat domains: MATH~\citep{Hendrycks2021Math}, LiveCodeBench~\citep{Jain2025Livecodebench}, and Alpaca~\citep{Taori2023Alpaca}, sampling 1,000 requests from each. We issue requests at concurrency levels $C\in\{8,16,32,64,128\}$ and report closed-loop throughput and average KV cache utilization during serving.

\begin{figure}[t]
\includegraphics[width=\linewidth]{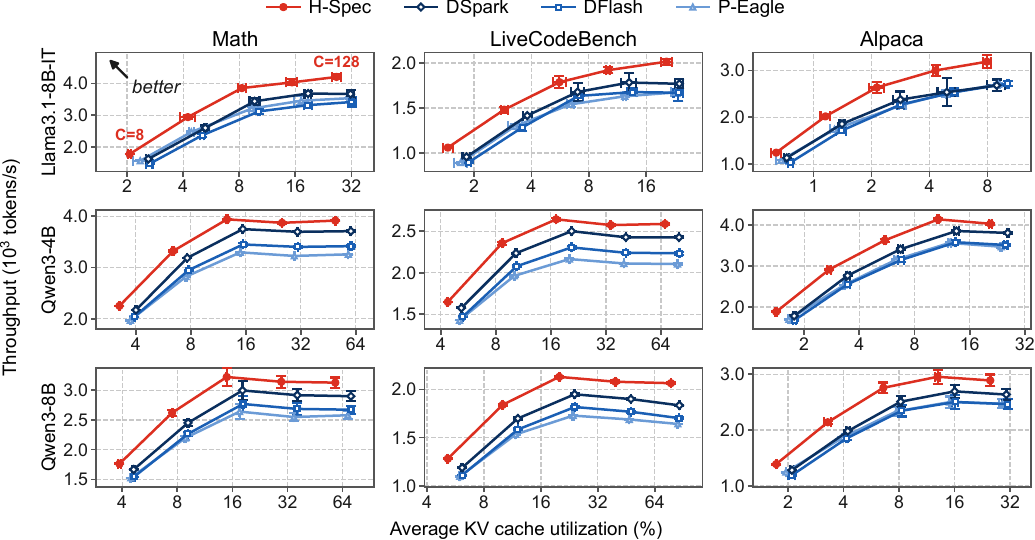}
\captionsetup{skip=5pt}
\caption{\label{fig:conc-tput-kv}\textbf{\sysname leads in throughput while maintaining the lowest KV cache utilization across concurrency levels.} Throughput and the average KV cache utilization during serving, under C=8--128 concurrent requests. Within a curve, markers from left to right correspond to increasing concurrency. Results are averaged from 3 repeats and error bars show standard deviation (st.d.).}
\end{figure}

\sysname achieves the highest throughput and lowest KV cache utilization at every evaluated concurrency level (Figure~\ref{fig:conc-tput-kv}). Relative to the strongest baseline, peak throughput improves by 13.0--17.3\%, 5.1--7.4\%, and 7.6--9.7\% across the three targets; average KV cache utilization is reduced by 4.3--24.6\% relative to the most memory-efficient baseline across all settings. \sysname's throughput advantage generally grows with concurrency. For example, on Qwen3-8B, the margin over the best baseline increases by 1.13--1.58$\times$ across three tasks from $C=8$ to $128$, consistent with the observation that baseline KV injection overhead scales with concurrency, which \sysname avoids.

\noindent\textbf{Preliminary comparison with DFlash-2.} Appendix~\secref{sec:eval-dflash2} compares \sysname against our DFlash-2 implementation following the authors' workflow~\citep{Inco2026Dflash2}, as no official training pipeline is available. On Qwen3-8B, \sysname achieves higher throughput and lower KV-cache utilization across all tasks and concurrency levels. Since DFlash-2 retains DFlash's block diffusion backbone, its drafter-side KV injection overhead still grows with concurrency (Figure~\ref{fig:kv-overhead-conc}).

\subsection{\label{sec:ablation}Ablation studies}

All ablation variants are trained from scratch using the same training recipe as the full \sysname. For the remaining analyses in the paper, we report unweighted average $\tau$ and Spd. from the eight tasks used in Table~\ref{tab:rhbench-mal-spd}, with per-task results omitted for brevity.

\noindent\textbf{Target-context source ablation.} We validate our central hypothesis: the last-token hidden states and the KVs of the target model provide complementary context for drafting. We therefore keep the \sysname architecture unchanged and ablate one context source at a time. In the \textit{Last-token-hidden-only} variant, attention is restricted to masked block positions, removing target KVs from the prefix context; in the \textit{KV-reuse-only} variant, the Mamba states are initialized with zeros, and weights for cross-layer fusion and projection of the last-token hidden states are removed as no longer needed.

\begin{figure*}[t]
    \centering
    \begin{subfigure}[t]{0.49\textwidth}
        \centering
        \includegraphics[width=\linewidth]{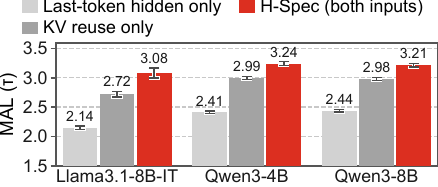}
        \captionsetup{skip=3pt}
        \caption{\textbf{Target-context source ablation.}}
        \label{fig:context-ablation}
    \end{subfigure}
    \hfill
    \begin{subfigure}[t]{0.49\textwidth}
        \centering
        \includegraphics[width=\linewidth]{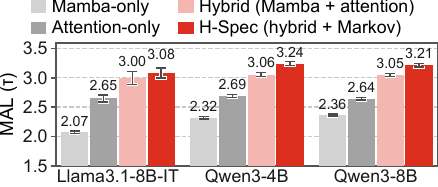}
        \captionsetup{skip=3pt}
        \caption{\textbf{Drafter architecture ablation.}}
        \label{fig:arch-ablation}
    \end{subfigure}

    \captionsetup{aboveskip=2pt,belowskip=1pt}
    \caption{\label{fig:ablations}
    \textbf{Ablation studies.}
    \ref{fig:context-ablation}: Hybrid context injection improves $\tau$ over either context source alone under a fixed drafter architecture.
    \ref{fig:arch-ablation}: The Mamba-attention architecture improves $\tau$ over either component alone under roughly matched parameter counts.
    Error bars are 95\% bootstrap CIs.}
\end{figure*}

Figure~\ref{fig:context-ablation} shows that using either target-context source alone reduces draft quality compared to full hybrid context injection. With the drafter architecture fixed, using only last-token hidden states or target KVs reduces average $\tau$ by 24.1--30.5\% and 7.1--11.7\%, respectively, across the three target models. These results confirm that the two context sources are complementary, not interchangeable.

\noindent\textbf{Drafter architecture ablation.} The study above ablates input context only. Here, we also change the drafter architecture: \textit{Mamba-only} and \textit{Attention-only} with only Mamba-MLP and attention-MLP layers, respectively; \textit{Hybrid} combines Mamba and attention. The first two have five layers to roughly match \sysname in total parameter count, with at most 5.2\% deviation across target models; detailed configurations are provided in Appendix~\secref{sec:ablation-settings}. Figure~\ref{fig:arch-ablation} shows that the hybrid backbone improves $\tau$ over Mamba-only by 28.9--45.1\% and over attention-only by 13.3--15.3\% under roughly matched parameter counts, showing that the combined architecture improves over either module alone.

\noindent\textbf{Drafter runtime scaling.} Throughput depends on both MAL and drafter runtimes. To isolate the effects, we compare the runtime of the hybrid Mamba-attention and block diffusion backbones from the Qwen3-8B drafters under a controlled setting: synthetic requests with 16 input and 2{,}048 output tokens, and all drafts accepted per step. Figure~\ref{fig:runtime-profiling} reports the hybrid backbone's relative speedup in forward-pass time (positive = hybrid faster), averaged over decoding steps across concurrency levels $C$ (\ref{fig:runtime-conc}), and across input lengths (binned by 256 tokens) at fixed $C=128$ (\ref{fig:runtime-input}). %

\begingroup
\setlength{\intextsep}{0pt}
\begin{wrapfigure}[15]{r}{0.493\linewidth}
\centering
\begin{subfigure}[t]{0.5387\linewidth}
  \includegraphics[width=\linewidth]{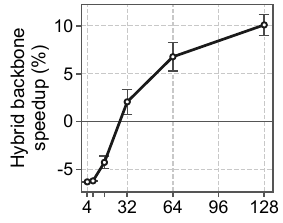}
  \captionsetup{skip=1.5pt,margin={0.1553\linewidth,0pt}}
  \caption{\label{fig:runtime-conc}Concurrency}
\end{subfigure}\hfill%
\begin{subfigure}[t]{0.4556\linewidth}
  \includegraphics[width=\linewidth]{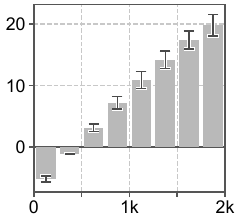}
  \captionsetup{skip=1.5pt}
  \caption{\label{fig:runtime-input}Input tokens}
\end{subfigure}
\captionsetup{skip=1.5pt}
\caption{\label{fig:runtime-profiling}\textbf{Hybrid vs. block diffusion runtime scaling.} Forward-pass speedup of the hybrid over the block-diffusion backbone across (a) concurrency and (b) input length at fixed $C=128$. Results are averaged from 5 waves per concurrency level and error bars show st.d. across waves.}
\end{wrapfigure}
At low concurrency ($C<32$), the hybrid backbone drafts more slowly than the block diffusion backbone, suggesting a deeper computational graph at matched parameter counts. However, it scales more favorably as $C$ increases by avoiding KV injection overhead. As input length increases, the hybrid backbone drafts more slowly initially but becomes faster beyond roughly 512 tokens. Mamba runtime is $\mathcal{O}(1)$ per step in input length, while attention runtime is $\mathcal{O}(N)$; fewer attention modules hence improve scaling to longer inputs. Overall, the hybrid backbone shows better runtime scaling for high-concurrency and long-context deployment.
\par
\endgroup

\subsection{\label{sec:sensitivity}Sensitivity analyses}

\begingroup
\setlength{\intextsep}{0pt}
\begin{wraptable}[10]{r}{0.49\textwidth}
\centering
\captionsetup{aboveskip=0pt,belowskip=0pt}
\caption{\label{tab:sampling-sensitivity}\textbf{Sensitivity to sampling parameters.} MAL under greedy/untruncated sampling. Statistical uncertainty reported in~\secref{sec:sensitivity-sampling-param-stat-uncert}.}
\vspace{2pt}
\fontsize{7.4pt}{8.6pt}\selectfont
\setlength{\tabcolsep}{1.8pt}
\renewcommand{\arraystretch}{1.02}
\newcommand{\tabnum}[1]{{\fontsize{8.2pt}{9.4pt}\selectfont #1}}
\setlength{\arrayrulewidth}{0.4pt}
\setlength{\aboverulesep}{0pt}\setlength{\belowrulesep}{0pt}
\begin{adjustbox}{width=\linewidth}
\begin{tabular}{@{}lcc|cc|cc@{}}
\toprule
\topstrut
& \multicolumn{2}{c|}{\textbf{Llama3.1-8B-IT}\botstrut}
& \multicolumn{2}{c|}{\textbf{Qwen3-4B}}
& \multicolumn{2}{c}{\textbf{Qwen3-8B}} \\
\cmidrule(lr){2-3} \cmidrule(lr){4-5} \cmidrule(lr){6-7}
\textbf{Method}\topstrut\botstrut
& Greedy & Untrunc.
& Greedy & Untrunc.
& Greedy & Untrunc. \\
\midrule
P-Eagle\topstrut
& \tabnum{2.93} & \tabnum{1.99}
& \tabnum{3.22} & \tabnum{2.75}
& \tabnum{3.09} & \tabnum{2.72} \\
DFlash
& \tabnum{2.57} & \tabnum{2.02}
& \tabnum{3.09} & \tabnum{2.66}
& \tabnum{2.91} & \tabnum{2.57} \\
DSpark
& \tabnum{2.86} & \tabnum{2.04}
& \tabnum{3.33} & \tabnum{2.83}
& \tabnum{3.15} & \tabnum{2.73} \\
\rowcolor{gray!12}
\sysname\botstrut
& \tabnum{\textbf{3.28}} & \tabnum{\textbf{2.20}}
& \tabnum{\textbf{3.49}} & \tabnum{\textbf{2.97}}
& \tabnum{\textbf{3.42}} & \tabnum{\textbf{2.96}} \\
\bottomrule
\end{tabular}
\end{adjustbox}
\vspace{-0.5\baselineskip}
\end{wraptable}
\textbf{Sensitivity to sampling parameters.} We validate that \sysname's lead in $\tau$ persists under extended sampling parameter configurations beyond the defaults used in Table~\ref{tab:rhbench-mal-spd}. As shown in Table~\ref{tab:sampling-sensitivity}, under greedy sampling (temp=0, topP/topK disabled), \sysname improves $\tau$ over the best baseline by 4.8--12.0\% across three targets. Under untruncated sampling (temp=1, topP=1, topK disabled), where all drafters decline, \sysname still leads in $\tau$ by 4.7--8.4\%.
\par
\endgroup

We report results for the Qwen3-8B target model in the remaining analyses; results for the other target models are in Appendix~\secref{sec:extended-analysis}.

\begin{figure}[H]
\centering
\begin{minipage}[t]{0.48\linewidth}
\vspace{0pt}
\centering
\captionsetup{skip=1.5pt}
\captionof{table}{\label{tab:draft-length}\textbf{Sensitivity to draft length.} MAL ($\tau$) and speedup for draft length $k\in\{3,5,7\}$. Statistical uncertainty reported in~\secref{sec:sensitivity-dl-stat-uncert}.}
\fontsize{7.8pt}{9.2pt}\selectfont
\setlength{\tabcolsep}{4.5pt}
\renewcommand{\arraystretch}{1.05}
\providecommand{\spx}{\kern0.05em{\fontsize{4.6pt}{5pt}\selectfont\texttimes}}
\providecommand{\topstrut}{\rule{0pt}{\dimexpr\arraystretch\ht\strutbox+2.80pt\relax}}
\providecommand{\botstrut}{\rule[-\dimexpr\arraystretch\dp\strutbox+1.72pt\relax]{0pt}{0pt}}
\setlength{\arrayrulewidth}{0.4pt}
\setlength{\aboverulesep}{0pt}\setlength{\belowrulesep}{0pt}
\begin{adjustbox}{width=\linewidth}
\begin{tabular}{lcc|cc|cc}
\toprule
\topstrut
& \multicolumn{2}{c|}{$k=3$\botstrut}
& \multicolumn{2}{c|}{$k=5$}
& \multicolumn{2}{c}{$k=7$} \\
\cmidrule(lr){2-3}
\cmidrule(lr){4-5}
\cmidrule(lr){6-7}
\textbf{Method}\topstrut\botstrut
& $\tau$ & \textit{Spd.}
& $\tau$ & \textit{Spd.}
& $\tau$ & \textit{Spd.} \\
\midrule
P-Eagle\topstrut
& 2.57 & \textit{2.05}\spx
& 2.84 & \textit{2.26}\spx
& 2.92 & \textit{2.30}\spx \\
DFlash
& 2.46 & \textit{1.99}\spx
& 2.68 & \textit{2.17}\spx
& 2.77 & \textit{2.28}\spx \\
DSpark
& 2.57 & \textit{2.07}\spx
& 2.90 & \textit{2.32}\spx
& 2.95 & \textit{2.40}\spx \\
\rowcolor{gray!12}
\sysname\botstrut
& \textbf{2.68} & {\bfseries\itshape 2.17}\spx
& \textbf{3.07} & {\bfseries\itshape 2.46}\spx
& \textbf{3.21} & {\bfseries\itshape 2.61}\spx \\
\bottomrule
\end{tabular}
\end{adjustbox}
\end{minipage}
\hfill
\begin{minipage}[t]{0.5\linewidth}
\vspace{0pt}
\centering
\includegraphics[width=\linewidth]{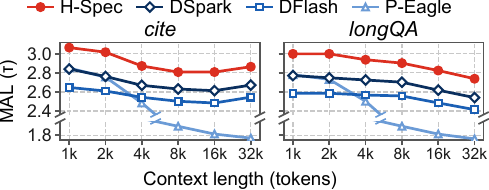}
\captionsetup{skip=1.5pt}
\captionof{figure}{\label{fig:longctx-mal}\textbf{Sensitivity to context length.} MAL ($\tau$) across context lengths on \textit{cite} and \textit{longQA} tasks. Statistical uncertainty reported in~\secref{sec:sensitivity-cl-stat-uncert}.}
\end{minipage}
\end{figure}

\noindent\textbf{Sensitivity to draft length.} We vary the inference-time draft length over $k\in\{3,5,7\}$ while using the same drafters trained with $k=7$. As shown in Table~\ref{tab:draft-length}, \sysname remains the best-performing drafter across all tested $k$ values. Across $k=3/5/7$, \sysname's $\tau$ margin over the best baseline is 4.3/5.8/8.7\%, and the speedup margin is 5.3/6.0/8.7\%, respectively. Speedup follows the same overall trend, showing that \sysname is robust to inference-time draft-length changes and generally gains a larger advantage at longer draft lengths.

\textbf{Sensitivity to context length.} Previous work shows draft quality can degrade on longer contexts~\citep{Eldenk2026Attentiondrift}. We therefore validate \sysname's robustness on two tasks from \textit{HELMET}~\citep{Yen2025Helmet}: \textit{cite} and \textit{longQA}, which support evaluation across multiple context-length bins. As Figure~\ref{fig:longctx-mal} shows, \sysname's $\tau$ margin over the best baseline remains 6.8--9.2\% across both tasks and all context lengths. From 1K to 32K, the margin changes only from +7.5 to +7.2\% on \textit{cite} and remains +7.8\% on \textit{longQA}, showing \sysname's advantage is sustained as context length increases.

\noindent\textbf{Transferability to finetuned targets.} We serve drafters unchanged on targets finetuned from their training target, reported in Appendix~\secref{sec:zero-shot}. Overall, \sysname retains the best $\tau$ and Spd. under target-model weight and/or chat template shifts, demonstrating zero-shot transferability without retraining.

\section{Related work, Limitations, Conclusion and Future Work}

\noindent\textbf{Related work.} Earlier drafters such as the Eagle series~\citep{Li2024Eagle, Li2024Eagle2, Li2025Eagle3} and ReDrafter~\citep{Cheng2024Recurrent} predict tokens autoregressively. P-Eagle~\citep{Hui2026Peagle} extends Eagle to parallel drafting by predicting multiple future positions from a shared learnable hidden state. DFlash~\citep{Chen2026Dflash} instead uses a block-diffusion drafter, conditioning on target-model hidden states by projecting them into a separate drafter-side KV cache. Subsequent drafters such as Domino~\citep{Huang2026Domino}, DSpark~\citep{Cheng2026Dspark}, DBlast~\citep{Karimi2026Dblast}, and DFlash-2~\citep{Inco2026Dflash2} retain the block-diffusion structure but improve inter-token dependencies through different auxiliary components. \sysname instead deviates from the conventional block-diffusion architecture by introducing hybrid target-context injection through a Mamba-attention architecture, eliminating the separate drafter-side KV cache.

An orthogonal direction augments the target model to predict multiple future tokens without a separate drafter, such as Medusa~\citep{Cai2024Medusa}, MTP~\citep{Gloeckle2024Better}, and Orthrus~\citep{Nguyen2026Orthrus}.
\sysname instead follows the standard speculative-decoding setup that leaves the target model entirely unchanged.

\noindent\textbf{Limitations.} \sysname requires its drafter-side K and V dimensions to match the target model's to reuse target KVs as-is, introducing an additional architectural constraint compared to hidden-state-based KV injection. The hybrid backbone drafts more slowly at lower concurrency or shorter input lengths, reflecting a deeper computational graph at matched parameter counts; at higher concurrency or longer inputs, this trend reverses as the hybrid architecture scales more favorably (Figure~\ref{fig:runtime-profiling}).

\noindent\textbf{Conclusion and future work.} In this work, we identify the drafter-side KV cache as an avoidable bottleneck in existing block diffusion drafters. We eliminate it with a novel hybrid target-context injection mechanism that combines last-token target hidden states with in-place target-KV reuse. Building on this mechanism, we propose \sysname, a parallel drafter that improves both single-request performance and concurrent-serving efficiency. More broadly, our results highlight system efficiency under concurrent serving as an important consideration in drafter architecture design. Future work will explore alternative target-layer selection strategies, partial target-KV head reuse to reduce drafter attention overhead, and additional causal correction schemes for the hybrid backbone.

\bibliography{references}

\appendix
\newpage

\startcontents[appendix]
\section*{Appendix Table of Contents}
\printcontents[appendix]{}{1}{}

\section{\label{sec:extended-background}Extended background on speculative decoding}

\noindent\textbf{Definition.} Speculative decoding~\citep{Leviathan2023Fast} losslessly accelerates the inference of a target LLM $\mathcal{M}_T$ with a smaller draft model $\mathcal{M}_D$. In each decoding iteration, given an input sequence $L$, $\mathcal{M}_D$ generates $k$ draft tokens $t_1\dots t_k$ with probabilities $P_D(t_i|L,t_1\dots t_{i-1})$. $\mathcal{M}_T$ then performs verification with a single forward pass over the concatenated sequence $L,t_1\dots t_{k}$, obtaining $P_T(t_i|L,t_1\dots t_{i-1})$ for $1\leq i\leq k + 1$.

A draft token $t_i$ is accepted if all preceding drafts are accepted and
\[\min(1,\frac{P_T(t_i|L,t_1\dots t_{i-1})}{P_D(t_i|L,t_1\dots t_{i-1})})\geq\alpha\]
where $\alpha\sim U(0,1)$ and $1 \leq i \leq k$.

\noindent\textbf{Target-generated ``bonus'' token.} Each verification produces one target-generated ``bonus'' token in addition to accepted drafts. If $t_i$ is the first rejected draft, all subsequent drafts are rejected, and $t_i$ is replaced by sampling from the corrected target distribution conditioned on $L,t_1\dots t_{i-1}$. If all $k$ drafts are accepted, a bonus token $t_{k+1}$ is sampled with probability $P_T(t_{k+1}|L,t_1\dots t_k)$.

\section{\label{sec:exp-config}Experimental settings}

\subsection{\label{sec:sys-config}\sysname architecture configuration}

This section details \sysname's architecture configuration. To ensure controlled comparisons, we match the attention, MLP, and, where applicable, Markov-head dimensions between \sysname and DFlash/DSpark, while closely matching their total trainable parameter counts. This helps ensure performance differences reflect the drafter architecture rather than mismatched drafter capacity or configurations.

\noindent\textbf{Number of drafter layers.} Available official DFlash and DSpark checkpoints for our evaluated targets use five drafter layers. We keep the attention and MLP dimensions matched to these baselines, but each \sysname layer additionally contains a Mamba module. To keep the total parameter count closely matched, \sysname therefore uses four layers: three full Mamba-attention-MLP layers and one partial Mamba-MLP layer.

\noindent\textbf{Attention.} \sysname matches the target model's attention configuration, including the Q, K, and V dimensions and the numbers of attention and KV heads, so that target-model KVs can be reused directly. All available official DFlash and DSpark checkpoints for our evaluated targets also match these attention configurations to their respective target models. Thus, although the baselines are not constrained by target-KV reuse, \sysname's target-matching requirement does not introduce an additional attention-configuration difference.

\noindent\textbf{MLP.} We set the MLP hidden dimension to match the target model, consistent with conventions in the official DFlash and DSpark checkpoints.

\noindent\textbf{Mamba.} We use a Mamba head dimension of 64, with the state dimension, number of groups, and convolution kernel size set to 4, and the expansion factor set to 1. We vary the number of Mamba heads across target models to minimize the remaining parameter-count gap with the baselines: 44 and 48 heads for Qwen3 4B and 8B, respectively, and 56 heads for Llama3.1-8B-IT. This keeps the difference in trainable parameters within 2.9\% compared to DSpark across all evaluated targets. We set $r=4{,}096$ for Llama3.1/Qwen3-8B, and $2{,}560$ for Qwen3-4B as the intermediate rank for cross-layer fusion (\secref{sec:mamba}), corresponding to the target model's hidden dimension. $d_S$, the flattened dimensionality of Mamba-2's recurrent state, equals the product of the number of Mamba heads, the per-head dimension, and the state dimension; the projected state is reshaped to [\text{n\_heads}, \text{head\_dim}, \text{d\_state}] before Mamba initialization.

\noindent\textbf{Markov rank.} We use a Markov rank of 256 for causal correction, matching DSpark.

\noindent\textbf{Target-layer mapping.} Existing block-diffusion drafters fuse and project hidden states from five evenly spaced target-model layers: layers 1, 8, 15, 22, and 30 for Llama3.1-8B-IT, and layers 1, 9, 17, 25, and 34 for Qwen3 4B and 8B. For a controlled comparison, \sysname draws target context from the same layers. The last-token hidden states from all five layers are fused and projected into the Mamba initial states, while the three attention modules reuse the KV caches from the last three selected layers: 15, 22, and 30 for Llama3.1-8B-IT, and 17, 25, and 34 for Qwen3. This ensures that \sysname's advantage does not result from using different target-model layer representations.

\subsection{\label{sec:baseline-config}Baseline configuration}

We report and justify any non-trivial configuration settings during baseline reproduction. Overall, we standardize each configuration choice across methods based on released checkpoints or prior literature, aiming to control the comparison rather than optimize for any individual method.

\noindent\textbf{Unified attention implementation.} Official baseline checkpoints use different attention implementations, such as Llama-style or Qwen3 attention. To ensure that comparisons reflect drafter architecture rather than attention implementation, we use Qwen3Attention for \sysname and all reproduced baselines. This follows the official DFlash implementation, which uses Qwen3 attention in its drafters for both Qwen3 and non-Qwen3 target models.

\noindent\textbf{Sliding-window attention (SWA).} We use SWA with a window size of 2{,}048 for \sysname and all reproduced baselines. SWA is used in the official DSpark and DFlash-2 releases~\citep{Inco2026Dflash2}, and has also been shown to improve long-context stability for EAGLE-style architectures~\citep{Eldenk2026Attentiondrift}. We therefore standardize on SWA for controlled attention-window configuration.

\noindent\textbf{Pruned drafter vocabulary.} All methods use the target model's LM-head weights. Following prior work on vocabulary pruning~\citep{Zhao2025Frspec, Goel2025Vocabtrim}, we restrict drafter prediction to the 32{,}000 most frequent target-vocabulary tokens observed in the training corpus. This pruning uniformly reduces drafting latency for all methods while maintaining lossless generation.

\noindent\textbf{Fixed-length verification.} We use fixed-length verification, where the target model verifies all drafted tokens in every iteration. Adaptive verification strategies, such as DSpark's confidence-scheduled verification~\citep{Cheng2026Dspark}, Speculative-Verification~\citep{Kim2026Speculative}, or D-cut~\citep{Liu2026Dcut}, operate at the verification stage and are orthogonal to the drafter architecture studied in this work. We therefore keep the verification procedure fixed across methods so that performance differences reflect the drafter itself rather than differences in verification policy.

\subsection{\label{sec:training-config}Training recipe}

We use a unified training recipe for \sysname and all reproduced baseline methods, while retaining method-specific training mechanisms where applicable:

\noindent\textbf{Training corpus.} We use 100K samples randomly drawn from Magpie~\citep{Xu2025Magpie} and Ultrachat~\citep{Ding2023Ultrachat} in a 60:40 ratio, proportional to the sizes of the original datasets. The scale of this corpus matches an official training corpus released by the DFlash authors.\footnote{\href{https://huggingface.co/datasets/jiamingshan/qwen3-4b-dflash-official100k-prepared}{https://huggingface.co/datasets/jiamingshan/qwen3-4b-dflash-official100k-prepared}} For Qwen3 4B and 8B, we use target-generated responses with thinking enabled rather than the original responses, allowing all drafters to better align with the thinking behavior.

\noindent\textbf{Preprocessing.} Every sample is tokenized into a training sequence and truncated to at most 8{,}192 tokens. We use Multipack sampling to pack these into 8{,}192-token sequences, with block-diagonal attention masks preventing attention across examples. Within each packed sequence, we randomly select up to 3{,}072 anchor positions, each corresponding to a block start for \sysname and the block diffusion baselines, i.e., DFlash and DSpark, or a chain start for P-Eagle.

\noindent\textbf{Optimizer settings and steps.} We use matched optimizer settings, learning-rate schedules, and optimizer steps for all drafters to ensure a controlled comparison. Specifically, we use AdamW with a peak learning rate of $6\times10^{-4}$, weight decay of 0.01, and gradient clipping at a global norm of 1.0, with 1\% linear warmup followed by half-cosine decay to zero. All methods train for 5 epochs on the same 100K samples across two GPUs with identical packing. With one packed sequence per GPU per optimization step, this gives a global batch size of two packed sequences (16{,}384 tokens), containing up to 6{,}144 anchors. This yields matched optimizer steps: 27{,}086 for Llama3.1-8B-IT and 104{,}032 for Qwen3 targets.

\noindent\textbf{Loss function and block-position weighting.} Following DSpark~\citep{Cheng2026Dspark}, we use a compound loss combining cross-entropy and total variation~\citep{Samarin2026Lk} for all drafters, weighted by 0.1 and 0.9, respectively. The Markov heads in DSpark and \sysname are trained with the same loss. For \sysname, DFlash, and DSpark, we apply block-position loss weighting as in DFlash and DSpark: $w_k=\exp(-(k-1)/\gamma)$, where $w_k$ is the loss weight at masked block position $1\leq k\leq7$ and $\gamma=4$. P-Eagle instead uses its own position-dependent sampling scheme: COD sampling retains a fraction $\max(0.6^d,0.2)$ of anchors at depth $0 \leq d \leq 6$~\citep{Hui2026Peagle}.

\begin{figure}[h]
\centering
\includegraphics[width=\linewidth]{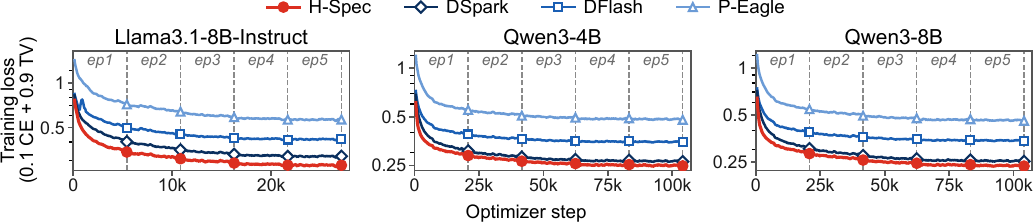}
\captionsetup{skip=1.5pt}
\caption{\label{fig:train-loss}\textbf{Training loss for \sysname and all reproduced baselines.} For each target model, all methods are trained for an identical number of optimizer steps, with vertical dotted lines marking epoch boundaries. Loss values are shown as moving averages over 1\% of the run for readability.}
\end{figure}

Figure~\ref{fig:train-loss} shows the training loss for all methods over three target models. All methods converge stably within the matched number of optimizer steps.

\subsection{\label{sec:ablation-settings}Settings for drafter architecture ablation}

This section provides detailed configurations for the three drafter variants studied in the architecture ablation in \secref{sec:ablation}. All three variants omit the Markov correction head: \textit{Mamba-only} uses Mamba-MLP layers with last-token hidden-state injection, \textit{Attention-only} uses attention-MLP layers with target-KV reuse, and \textit{Hybrid} combines Mamba and attention. The full \sysname adds the Markov head on top of the hybrid backbone.

For the \textit{Mamba-only} and \textit{Attention-only} variants, we keep the dimensions of the Mamba/attention and MLP modules unchanged from the full \sysname, and increase both to five layers for parameter matching, with at most a 5.2\% difference in total parameter count. In the full \sysname, Mamba initial-state injection uses last-token hidden states from five selected target layers, while the three attention modules reuse KVs from three of the five selected layers. In \textit{Attention-only}, the two additional attention-MLP layers reuse KVs from the remaining two target layers, keeping the overall set of selected target layers unchanged.

\section{\label{sec:stat-uncert}Statistical uncertainty analysis in evaluation results}

We report 95\% bootstrap confidence intervals (CIs) to quantify the statistical uncertainty in MAL ($\tau$) and ITL speedup, and use paired bootstrap analyses to assess whether \sysname's improvements over the baselines are statistically significant.

For single-task $\tau$ and Spd., we estimate 95\% CIs using 10{,}000 bootstrap resamples of the task's requests: each resample is drawn with replacement from the original requests, and we recompute $\tau$ and Spd. on the resampled set. For results averaged over the eight tasks, we independently resample the requests within each task, recompute the task-level $\tau$ and Spd., and then average each metric across the eight tasks. We repeat this procedure 10{,}000 times to obtain the bootstrap distribution of the task-average $\tau$ and Spd., from which we estimate their 95\% CIs.

\subsection{\label{sec:single-request-stat-uncert}Single-request MAL and speedup}

Complementing Table~\ref{tab:rhbench-mal-spd}, we assess whether \sysname's improvement over baselines is statistically significant, both when averaged across eight tasks and separately per task.

\noindent\textbf{Average across all tasks.} We conduct paired bootstrap analyses between \sysname and DSpark, which is the best baseline in task-average $\tau$ and Spd. for all three target models.  The 95\% CIs of \sysname's relative improvement over DSpark in $\tau$ are [+9.0\%, +17.5\%], [+4.3\%, +5.6\%], [+8.0\%, +9.4\%], and in Spd. are [+10.2\%, +15.1\%], [+4.7\%, +5.9\%], [+7.9\%, +9.4\%], across Llama3.1, Qwen3 4B and 8B target models, respectively. All CIs lie entirely above 0, demonstrating the statistical significance of \sysname's gains.

\noindent\textbf{Per task/model setting.} Across eight tasks and three target models, \sysname achieves the best $\tau$ in 22/24 settings and the best Spd. in 23/24 settings. We conduct paired bootstrap analyses for each of the 45 (task, target model, metric) settings where \sysname leads, comparing \sysname with the corresponding best baseline in that setting. In 42/45 cases, the 95\% CI lies above zero, showing that \sysname's lead is statistically significant. In the remaining three cases (Spd. in Math and Chat on Llama3.1, and $\tau$ in Transl. on Qwen3-4B), the lead over the best baseline is not statistically significant as the CI includes zero, but the improvement remains significant compared with the second-best baseline.

\subsection{\label{sec:sensitivity-sampling-param-stat-uncert}Sensitivity to sampling parameters}

Complementing Table~\ref{tab:sampling-sensitivity}, we conduct paired bootstrap analyses between \sysname and the best baseline (P-Eagle for Llama3.1/Greedy, DSpark in all remaining settings) under all sampling configurations and target model settings. Under greedy sampling, the 95\% CI of \sysname's $\tau$ improvement over the best baseline is [+8.1\%, +15.3\%], [+3.7\%, +5.8\%] and [+7.6\%, +9.4\%], for Llama3.1, Qwen3 4B and 8B target models, respectively. For untruncated sampling, the 95\% CIs are [+0.6\%, +18.6\%], [+4.1\%, +5.2\%] and [+7.5\%, +9.5\%]. All CIs lie entirely above zero, demonstrating that \sysname's gain in $\tau$ is statistically significant across multiple sampling parameter settings.

\subsection{\label{sec:sensitivity-dl-stat-uncert}Sensitivity to inference-time draft length}

To complement Table~\ref{tab:draft-length}, we run paired bootstrap analyses between \sysname and DSpark, the best baseline across all inference-time draft-length settings. For $k=3/5/7$, the 95\% bootstrap CI for \sysname's $\tau$ margin over DSpark is [+3.8\%, +4.8\%], [+5.1\%, +6.4\%], and [+8.0\%, +9.4\%]; for speedup, the CIs are [+4.8\%, +5.8\%], [+5.5\%, +6.5\%], and [+7.9\%, +9.4\%]. All paired CIs lie above zero, showing that \sysname's lead is statistically significant across various inference-time draft length settings.

\subsection{\label{sec:sensitivity-cl-stat-uncert}Sensitivity to context length}

To complement Figure~\ref{fig:longctx-mal}, we conduct paired bootstrap analyses between \sysname and the best baseline at every (task, context length bin) setting; the best baseline is P-Eagle for context length 1K on both tasks, and DSpark in remaining settings. The 95\% CI is above zero for every setting, with the tightest margin being [+5.8\%, +8.6\%] on \textit{cite} at 32K context. This result shows that \sysname's hybrid context injection remains effective across context lengths, suggesting robustness to the attention drift~\citep{Eldenk2026Attentiondrift}.

\section{Extended evaluation}

\subsection{\label{sec:deepspec-eval}Single-request evaluation on DeepSpec tasks}

Complementing~\secref{sec:single-request-eval}, we provide additional mean accepted length ($\tau$) and ITL speedup evaluations on the nine-task benchmark suite released by the DSpark authors, which we refer to as the DeepSpec suite. These tasks are used in the original DFlash and DSpark papers and improve the coverage of our evaluation.

\begin{table*}[h]
\centering
\captionsetup{skip=1.5pt}
\caption{\label{tab:deepspec-mal-spd}\textbf{\sysname leads in draft acceptance and speedup on the DeepSpec evaluation suite.} Mean accepted length ($\tau$) and batch-size-1 ITL speedup (Spd.) relative to no-speculative-decoding over nine tasks. Paired bootstrap 95\% CIs for \sysname's relative gains over the best baseline lie above zero for every task-target-metric setting, indicating statistical significance.}
\setlength{\tabcolsep}{1.95pt}
\renewcommand{\arraystretch}{1.10}
\fontsize{8}{9.6}\selectfont
\providecommand{\spx}{\kern0.05em{\fontsize{4.6pt}{5pt}\selectfont\texttimes}}
\setlength{\arrayrulewidth}{0.4pt}\arrayrulecolor{black!42}
\begin{adjustbox}{width=\linewidth}
\begin{tabular}{l cc|cc|cc|cc|cc|cc|cc|cc|cc}
\arrayrulecolor{black}\specialrule{\heavyrulewidth}{0pt}{0pt}\arrayrulecolor{black!42}
\rule{0pt}{2.6ex} & \multicolumn{6}{c|}{\textbf{Math}} & \multicolumn{6}{c|}{\textbf{Code}} & \multicolumn{6}{c}{\textbf{Chat}} \\
\arrayrulecolor{black!42}\cline{2-7}\cline{8-13}\cline{14-19}
\rule{0pt}{2.5ex}\rule[-0.9ex]{0pt}{0pt} & \multicolumn{2}{c|}{{\scriptsize gsm8k}} & \multicolumn{2}{c|}{{\scriptsize math500}} & \multicolumn{2}{c|}{{\scriptsize aime25}} & \multicolumn{2}{c|}{{\scriptsize livecodebench}} & \multicolumn{2}{c|}{{\scriptsize mbpp}} & \multicolumn{2}{c|}{{\scriptsize humaneval}} & \multicolumn{2}{c|}{{\scriptsize alpaca}} & \multicolumn{2}{c|}{{\scriptsize arena-hard-v2}} & \multicolumn{2}{c}{{\scriptsize mt-bench}} \\
\arrayrulecolor{black!60}\specialrule{0.7pt}{0pt}{0pt}\arrayrulecolor{black!42}
\rule{0pt}{2.7ex}{\scshape\color{black!62}L-8B-IT}\rule[-0.8ex]{0pt}{0pt} & {\scriptsize\color{black!62}$\tau$} & {\scriptsize\color{black!62}\textit{Spd.}} & {\scriptsize\color{black!62}$\tau$} & {\scriptsize\color{black!62}\textit{Spd.}} & {\scriptsize\color{black!62}$\tau$} & {\scriptsize\color{black!62}\textit{Spd.}} & {\scriptsize\color{black!62}$\tau$} & {\scriptsize\color{black!62}\textit{Spd.}} & {\scriptsize\color{black!62}$\tau$} & {\scriptsize\color{black!62}\textit{Spd.}} & {\scriptsize\color{black!62}$\tau$} & {\scriptsize\color{black!62}\textit{Spd.}} & {\scriptsize\color{black!62}$\tau$} & {\scriptsize\color{black!62}\textit{Spd.}} & {\scriptsize\color{black!62}$\tau$} & {\scriptsize\color{black!62}\textit{Spd.}} & {\scriptsize\color{black!62}$\tau$} & {\scriptsize\color{black!62}\textit{Spd.}} \\
\arrayrulecolor{black!30}\specialrule{0.3pt}{0pt}{0pt}\arrayrulecolor{black!42}
\rule{0pt}{2.6ex}P-Eagle & 3.21 & \textit{2.66}\spx & 3.96 & \textit{3.19}\spx & 4.05 & \textit{3.16}\spx & 2.75 & \textit{2.25}\spx & 3.53 & \textit{2.90}\spx & 3.50 & \textit{2.87}\spx & 2.57 & \textit{2.09}\spx & 2.13 & \textit{1.68}\spx & 2.61 & \textit{2.17}\spx \\
DFlash & 3.11 & \textit{2.65}\spx & 3.78 & \textit{3.14}\spx & 3.86 & \textit{3.03}\spx & 2.67 & \textit{2.23}\spx & 3.46 & \textit{2.91}\spx & 3.26 & \textit{2.72}\spx & 2.40 & \textit{2.02}\spx & 2.08 & \textit{1.63}\spx & 2.45 & \textit{2.11}\spx \\
DSpark & 3.42 & \textit{2.83}\spx & 4.17 & \textit{3.37}\spx & 4.18 & \textit{3.36}\spx & 2.90 & \textit{2.37}\spx & 3.86 & \textit{3.03}\spx & 3.67 & \textit{3.05}\spx & 2.70 & \textit{2.22}\spx & 2.08 & \textit{1.65}\spx & 2.92 & \textit{2.28}\spx \\
\rowcolor{gray!12}\sysname & \textbf{4.02} & {\bfseries\itshape 3.35}\spx & \textbf{4.64} & {\bfseries\itshape 3.78}\spx & \textbf{4.78} & {\bfseries\itshape 3.74}\spx & \textbf{3.33} & {\bfseries\itshape 2.66}\spx & \textbf{4.21} & {\bfseries\itshape 3.51}\spx & \textbf{4.20} & {\bfseries\itshape 3.39}\spx & \textbf{3.01} & {\bfseries\itshape 2.48}\spx & \textbf{2.31} & {\bfseries\itshape 1.90}\spx & \textbf{3.14} & {\bfseries\itshape 2.54}\spx\rule[-0.8ex]{0pt}{0pt} \\
\arrayrulecolor{black!60}\specialrule{0.7pt}{0pt}{0pt}\arrayrulecolor{black!42}
\rule{0pt}{2.7ex}{\scshape\color{black!62}Q-4B}\rule[-0.8ex]{0pt}{0pt} & {\scriptsize\color{black!62}$\tau$} & {\scriptsize\color{black!62}\textit{Spd.}} & {\scriptsize\color{black!62}$\tau$} & {\scriptsize\color{black!62}\textit{Spd.}} & {\scriptsize\color{black!62}$\tau$} & {\scriptsize\color{black!62}\textit{Spd.}} & {\scriptsize\color{black!62}$\tau$} & {\scriptsize\color{black!62}\textit{Spd.}} & {\scriptsize\color{black!62}$\tau$} & {\scriptsize\color{black!62}\textit{Spd.}} & {\scriptsize\color{black!62}$\tau$} & {\scriptsize\color{black!62}\textit{Spd.}} & {\scriptsize\color{black!62}$\tau$} & {\scriptsize\color{black!62}\textit{Spd.}} & {\scriptsize\color{black!62}$\tau$} & {\scriptsize\color{black!62}\textit{Spd.}} & {\scriptsize\color{black!62}$\tau$} & {\scriptsize\color{black!62}\textit{Spd.}} \\
\arrayrulecolor{black!30}\specialrule{0.3pt}{0pt}{0pt}\arrayrulecolor{black!42}
\rule{0pt}{2.6ex}P-Eagle & 3.76 & \textit{2.80}\spx & 3.83 & \textit{2.81}\spx & 3.52 & \textit{2.54}\spx & 3.07 & \textit{2.17}\spx & 3.35 & \textit{2.51}\spx & 3.37 & \textit{2.49}\spx & 2.67 & \textit{2.03}\spx & 2.27 & \textit{1.65}\spx & 2.95 & \textit{2.19}\spx \\
DFlash & 3.70 & \textit{2.96}\spx & 3.79 & \textit{3.00}\spx & 3.50 & \textit{2.74}\spx & 2.98 & \textit{2.27}\spx & 3.29 & \textit{2.65}\spx & 3.29 & \textit{2.59}\spx & 2.56 & \textit{2.11}\spx & 2.19 & \textit{1.71}\spx & 2.85 & \textit{2.27}\spx \\
DSpark & 4.01 & \textit{3.13}\spx & 4.12 & \textit{3.18}\spx & 3.78 & \textit{2.85}\spx & 3.26 & \textit{2.41}\spx & 3.60 & \textit{2.84}\spx & 3.62 & \textit{2.77}\spx & 2.79 & \textit{2.22}\spx & 2.31 & \textit{1.74}\spx & 3.09 & \textit{2.42}\spx \\
\rowcolor{gray!12}\sysname & \textbf{4.14} & {\bfseries\itshape 3.23}\spx & \textbf{4.27} & {\bfseries\itshape 3.29}\spx & \textbf{3.91} & {\bfseries\itshape 2.97}\spx & \textbf{3.40} & {\bfseries\itshape 2.53}\spx & \textbf{3.76} & {\bfseries\itshape 2.96}\spx & \textbf{3.76} & {\bfseries\itshape 2.89}\spx & \textbf{2.92} & {\bfseries\itshape 2.35}\spx & \textbf{2.38} & {\bfseries\itshape 1.82}\spx & \textbf{3.21} & {\bfseries\itshape 2.53}\spx\rule[-0.8ex]{0pt}{0pt} \\
\arrayrulecolor{black!60}\specialrule{0.7pt}{0pt}{0pt}\arrayrulecolor{black!42}
\rule{0pt}{2.7ex}{\scshape\color{black!62}Q-8B}\rule[-0.8ex]{0pt}{0pt} & {\scriptsize\color{black!62}$\tau$} & {\scriptsize\color{black!62}\textit{Spd.}} & {\scriptsize\color{black!62}$\tau$} & {\scriptsize\color{black!62}\textit{Spd.}} & {\scriptsize\color{black!62}$\tau$} & {\scriptsize\color{black!62}\textit{Spd.}} & {\scriptsize\color{black!62}$\tau$} & {\scriptsize\color{black!62}\textit{Spd.}} & {\scriptsize\color{black!62}$\tau$} & {\scriptsize\color{black!62}\textit{Spd.}} & {\scriptsize\color{black!62}$\tau$} & {\scriptsize\color{black!62}\textit{Spd.}} & {\scriptsize\color{black!62}$\tau$} & {\scriptsize\color{black!62}\textit{Spd.}} & {\scriptsize\color{black!62}$\tau$} & {\scriptsize\color{black!62}\textit{Spd.}} & {\scriptsize\color{black!62}$\tau$} & {\scriptsize\color{black!62}\textit{Spd.}} \\
\arrayrulecolor{black!30}\specialrule{0.3pt}{0pt}{0pt}\arrayrulecolor{black!42}
\rule{0pt}{2.6ex}P-Eagle & 3.71 & \textit{2.98}\spx & 3.83 & \textit{3.04}\spx & 3.57 & \textit{2.80}\spx & 3.02 & \textit{2.34}\spx & 3.27 & \textit{2.62}\spx & 3.33 & \textit{2.63}\spx & 2.68 & \textit{2.16}\spx & 2.25 & \textit{1.77}\spx & 2.94 & \textit{2.32}\spx \\
DFlash & 3.58 & \textit{2.98}\spx & 3.75 & \textit{3.07}\spx & 3.47 & \textit{2.79}\spx & 2.92 & \textit{2.34}\spx & 3.18 & \textit{2.64}\spx & 3.18 & \textit{2.63}\spx & 2.54 & \textit{2.13}\spx & 2.14 & \textit{1.76}\spx & 2.78 & \textit{2.31}\spx \\
DSpark & 3.90 & \textit{3.19}\spx & 4.09 & \textit{3.31}\spx & 3.78 & \textit{3.02}\spx & 3.15 & \textit{2.49}\spx & 3.49 & \textit{2.84}\spx & 3.50 & \textit{2.83}\spx & 2.73 & \textit{2.27}\spx & 2.25 & \textit{1.81}\spx & 2.99 & \textit{2.44}\spx \\
\rowcolor{gray!12}\sysname & \textbf{4.14} & {\bfseries\itshape 3.40}\spx & \textbf{4.32} & {\bfseries\itshape 3.52}\spx & \textbf{4.00} & {\bfseries\itshape 3.28}\spx & \textbf{3.37} & {\bfseries\itshape 2.69}\spx & \textbf{3.69} & {\bfseries\itshape 3.03}\spx & \textbf{3.74} & {\bfseries\itshape 3.05}\spx & \textbf{2.93} & {\bfseries\itshape 2.45}\spx & \textbf{2.39} & {\bfseries\itshape 1.92}\spx & \textbf{3.22} & {\bfseries\itshape 2.68}\spx\rule[-0.8ex]{0pt}{0pt} \\
\arrayrulecolor{black}\specialrule{\heavyrulewidth}{0pt}{0pt}\arrayrulecolor{black!42}
\end{tabular}
\end{adjustbox}\arrayrulecolor{black}
\end{table*}

As shown in Table~\ref{tab:deepspec-mal-spd}, \sysname continues to lead in draft acceptance and batch-size-1 ITL speedup. Averaged over 9 tasks, \sysname improves in $\tau$ over the best baseline by 12.5\%, 3.8\%, and 6.4\% for the Llama, Qwen3 4B, and 8B target models, respectively. For speedup, the margins over the best baseline are 13.1\%, 4.3\%, and 7.5\% across three targets. Importantly, \sysname leads in every task for every evaluated target model, with the pairwise 95\% bootstrap CIs sitting above zero in all cases. These results further demonstrate \sysname's consistent gains across diverse domains.

\subsection{\label{sec:eval-dflash2}Preliminary comparison with DFlash-2}

We provide a preliminary comparison with DFlash-2~\citep{Inco2026Dflash2}, the most recent follow-up work from the DFlash team. DFlash-2 keeps the block diffusion drafter backbone and the KV injection mechanism (Figure~\ref{fig:kv-injection}) as in the original DFlash, but with two additions: (1) a two-tap convolution kernel before and after every attention and MLP module that mixes each position's hidden representation with that of its preceding position, and (2) a path selector that scores adjacent pairs among the top-16 candidates at each draft position and selects a coherent drafted sequence. Together, these additions better capture inter-token dependencies and provide a competitive alternative to existing causal correction mechanisms.

We reproduce DFlash-2 under the matched training recipes used in~\secref{sec:evaluation} on the Qwen3-8B target model. Following~\secref{sec:serving-eval}, we report decoding throughput (output tokens per second) and average KV cache utilization (\%) during request serving under concurrency levels $C\in\{8, 16, 32, 64, 128\}$, on the MATH, LiveCodeBench, and Alpaca workloads.

\noindent\textbf{Disclaimer:} We implemented DFlash-2's training and inference pipelines using the Speculators~\citep{Redhat2025Speculators} and vLLM~\citep{Kwon2023Efficient} frameworks to the best of our ability. Overall, our DFlash-2 implementation outperforms both DFlash and DSpark under a unified training recipe, consistent with the trend reported by the DFlash-2 authors.

\begin{figure}[h]
\includegraphics[width=\textwidth]{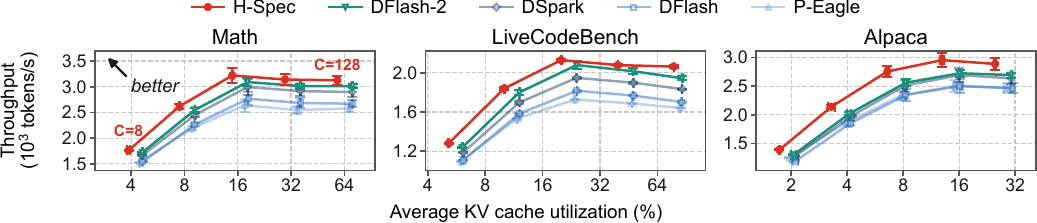}
\captionsetup{skip=1.5pt}
\caption{\label{fig:dflash2-kv-tput}\noindent\textbf{\sysname achieves higher throughput while maintaining lower KV cache utilization than DFlash-2 across concurrency levels.} Throughput and the average KV cache utilization during serving, under C=8--128 concurrent requests. Markers represent rising concurrency levels from left to right per curve. Results are averaged from three repeats, and error bars show st.d. across repeated runs.}
\end{figure}

As shown in Figure~\ref{fig:dflash2-kv-tput}, \sysname improves maximum throughput over DFlash-2 by 4.2\% on MATH, 2.4\% on LiveCodeBench, and 8.4\% on Alpaca, while reducing average KV cache utilization by 9.7\% to 20.2\% across all concurrency and task settings. DFlash-2 trails \sysname in maximum throughput by the smallest margin among all baselines, reflecting its improved drafting quality over DSpark. However, DFlash-2 does not narrow the KV cache utilization gap, since its block diffusion backbone and KV injection mechanism remain unchanged.

We conclude that \sysname and DFlash-2 make largely orthogonal contributions: \sysname proposes novel target-context injection via a hybrid Mamba-attention backbone, while DFlash-2 introduces convolution and a path selector as additional components to a parallel drafter backbone to capture inter-token dependencies. These components could be incorporated into our design: the convolution can be added to the hybrid backbone, while the path selector can replace the Markov head as the causal correction mechanism. We leave the evaluation and integration of these components as an important direction for future work.

\subsection{\label{sec:extended-analysis}Extended sensitivity analysis}

Complementing~\secref{sec:sensitivity}, we report analysis experiments on additional target models: Llama3.1-8B-IT and Qwen3-4B. Overall, our analysis findings hold across target models.

\begin{figure}[t]
\centering
\begin{minipage}[t]{0.48\linewidth}
\vspace{0pt}
\centering
\captionsetup{skip=1.5pt}
\captionof{table}{\label{tab:draft-length-extended}
\textbf{Sensitivity to inference draft length.}
MAL ($\tau$) and batch-size-1 speedup on Llama3.1-8B-IT and Qwen3-4B for draft lengths $k\in\{3,5,7\}$.}
\fontsize{7.8pt}{9.2pt}\selectfont
\setlength{\tabcolsep}{4.0pt}
\renewcommand{\arraystretch}{1.05}
\providecommand{\spx}{\kern0.05em{\fontsize{4.6pt}{5pt}\selectfont\texttimes}}
\setlength{\arrayrulewidth}{0.4pt}\arrayrulecolor{black!42}
\begin{adjustbox}{width=\linewidth}
\begin{tabular}{l cc|cc|cc}
\arrayrulecolor{black}\specialrule{\heavyrulewidth}{0pt}{0pt}\arrayrulecolor{black!42}
\rule{0pt}{2.6ex}\rule[-0.9ex]{0pt}{0pt} & \multicolumn{2}{c|}{$k=3$} & \multicolumn{2}{c|}{$k=5$} & \multicolumn{2}{c}{$k=7$} \\
\arrayrulecolor{black!60}\specialrule{0.7pt}{0pt}{0pt}\arrayrulecolor{black!42}
\rule{0pt}{2.7ex}{\scshape\color{black!62}L-8B-IT}\rule[-0.8ex]{0pt}{0pt} & {\scriptsize\color{black!62}$\tau$} & {\scriptsize\color{black!62}\textit{Spd.}} & {\scriptsize\color{black!62}$\tau$} & {\scriptsize\color{black!62}\textit{Spd.}} & {\scriptsize\color{black!62}$\tau$} & {\scriptsize\color{black!62}\textit{Spd.}} \\
\arrayrulecolor{black!30}\specialrule{0.3pt}{0pt}{0pt}\arrayrulecolor{black!42}
\rule{0pt}{2.6ex}P-Eagle & 2.40 & \textit{1.99}\spx & 2.68 & \textit{2.18}\spx & 2.69 & \textit{2.16}\spx \\
DFlash & 2.23 & \textit{1.85}\spx & 2.38 & \textit{1.93}\spx & 2.47 & \textit{2.03}\spx \\
DSpark & 2.32 & \textit{1.91}\spx & 2.64 & \textit{2.14}\spx & 2.72 & \textit{2.23}\spx \\
\rowcolor{gray!12}\sysname & \textbf{2.55} & {\bfseries\itshape 2.12}\spx & \textbf{2.98} & {\bfseries\itshape 2.43}\spx & \textbf{3.08} & {\bfseries\itshape 2.51}\spx\rule[-0.8ex]{0pt}{0pt} \\
\arrayrulecolor{black!60}\specialrule{0.7pt}{0pt}{0pt}\arrayrulecolor{black!42}
\rule{0pt}{2.7ex}{\scshape\color{black!62}Q-4B}\rule[-0.8ex]{0pt}{0pt} & {\scriptsize\color{black!62}$\tau$} & {\scriptsize\color{black!62}\textit{Spd.}} & {\scriptsize\color{black!62}$\tau$} & {\scriptsize\color{black!62}\textit{Spd.}} & {\scriptsize\color{black!62}$\tau$} & {\scriptsize\color{black!62}\textit{Spd.}} \\
\arrayrulecolor{black!30}\specialrule{0.3pt}{0pt}{0pt}\arrayrulecolor{black!42}
\rule{0pt}{2.6ex}P-Eagle & 2.59 & \textit{1.92}\spx & 2.87 & \textit{2.12}\spx & 2.99 & \textit{2.18}\spx \\
DFlash & 2.51 & \textit{1.92}\spx & 2.79 & \textit{2.11}\spx & 2.86 & \textit{2.26}\spx \\
DSpark & 2.60 & \textit{1.97}\spx & 2.93 & \textit{2.20}\spx & 3.09 & \textit{2.37}\spx \\
\rowcolor{gray!12}\sysname & \textbf{2.69} & {\bfseries\itshape 2.05}\spx & \textbf{3.09} & {\bfseries\itshape 2.32}\spx & \textbf{3.24} & {\bfseries\itshape 2.50}\spx\rule[-0.8ex]{0pt}{0pt} \\
\arrayrulecolor{black}\specialrule{\heavyrulewidth}{0pt}{0pt}\arrayrulecolor{black!42}
\end{tabular}
\end{adjustbox}\arrayrulecolor{black}
\end{minipage}
\hfill
\begin{minipage}[t]{0.50\linewidth}
\vspace{0pt}
\centering
\includegraphics[width=\linewidth]{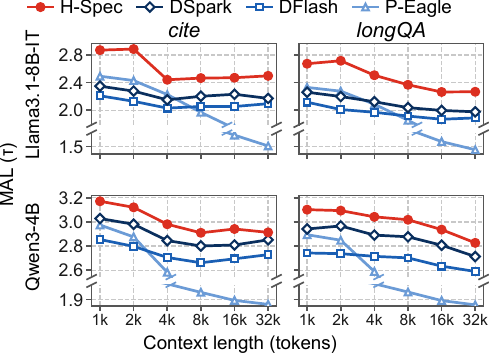}
\captionsetup{skip=1.5pt}
\captionof{figure}{\label{fig:longctx-mal-appendix}
\textbf{Sensitivity to context length.}
MAL ($\tau$) as a function of context length on \textit{cite} and \textit{longQA} for Llama3.1-8B-IT and Qwen3-4B.}
\end{minipage}
\end{figure}

\noindent\textbf{Sensitivity to draft length.} Table~\ref{tab:draft-length-extended} shows that \sysname continues to lead in $\tau$ and speedup across inference-time draft length for both target models. The margin in $\tau$ and speedup over the best baseline roughly increases with draft length, consistent with our observation on the Qwen3-8B target.

\noindent\textbf{Long-context generalization.} As shown in Figure~\ref{fig:longctx-mal-appendix}, \sysname achieves the best $\tau$ across context lengths on both \textit{HELMET} tasks for the additional targets. This shows that hybrid context injection provides strong, sustained target conditioning as context length increases.

\subsection{\label{sec:zero-shot}Zero-shot transfer to finetuned target models}

To evaluate the robustness of \sysname's gains under target-model shifts, we serve drafters under zero-shot transfer on additional target models finetuned from their original training target. We report $\tau$ and Spd. averaged over eight augmented Spec-Bench tasks.

\begin{table*}[h]
\centering
\captionsetup{skip=1.5pt}
\caption{\label{tab:finetuned-transfer}\textbf{\sysname remains the best drafter under zero-shot transfer.} Drafters trained for the base Llama3.1-8B-IT, Qwen3-4B, and Qwen3-8B target models are evaluated on two fine-tuned targets each. $\dagger$~indicates that the chat template changes in addition to the model weights.}
\fontsize{8.8pt}{10.2pt}\selectfont
\renewcommand{\arraystretch}{1.08}
\setlength{\tabcolsep}{6.25pt}
\providecommand{\spx}{\kern0.05em{\fontsize{4.6pt}{5pt}\selectfont\texttimes}}
\setlength{\arrayrulewidth}{0.4pt}\arrayrulecolor{black!42}
\begin{adjustbox}{width=\linewidth}
\begin{tabular}{l cc|cc|cc|cc|cc|cc}
\arrayrulecolor{black}\specialrule{\heavyrulewidth}{0pt}{0pt}\arrayrulecolor{black!42}
\rule{0pt}{2.6ex} & \multicolumn{4}{c|}{\textbf{Llama3.1-8B-IT}} & \multicolumn{4}{c|}{\textbf{Qwen3-4B}} & \multicolumn{4}{c}{\textbf{Qwen3-8B}} \\
\arrayrulecolor{black!42}\cline{2-5}\cline{6-9}\cline{10-13}
\rule{0pt}{2.5ex} & \multicolumn{2}{c|}{Selene-1} & \multicolumn{2}{c|}{R1-Distill$^\dagger$} & \multicolumn{2}{c|}{SFT-Sci} & \multicolumn{2}{c|}{Jan-Nano$^\dagger$} & \multicolumn{2}{c|}{II-Medical} & \multicolumn{2}{c}{R1-0528$^\dagger$} \\
\textbf{Method}\rule[-0.8ex]{0pt}{0pt} & {\scriptsize\color{black!62}$\tau$} & {\scriptsize\color{black!62}\textit{Spd.}} & {\scriptsize\color{black!62}$\tau$} & {\scriptsize\color{black!62}\textit{Spd.}} & {\scriptsize\color{black!62}$\tau$} & {\scriptsize\color{black!62}\textit{Spd.}} & {\scriptsize\color{black!62}$\tau$} & {\scriptsize\color{black!62}\textit{Spd.}} & {\scriptsize\color{black!62}$\tau$} & {\scriptsize\color{black!62}\textit{Spd.}} & {\scriptsize\color{black!62}$\tau$} & {\scriptsize\color{black!62}\textit{Spd.}} \\
\arrayrulecolor{black!30}\specialrule{0.3pt}{0pt}{0pt}\arrayrulecolor{black!42}
\rule{0pt}{2.6ex}P-Eagle & 2.71 & \textit{2.08}\spx & 1.88 & \textit{1.49}\spx & 2.55 & \textit{1.89}\spx & 2.63 & \textit{2.16}\spx & 2.54 & \textit{2.08}\spx & 1.98 & \textit{1.63}\spx \\
DFlash & 2.44 & \textit{2.00}\spx & 1.72 & \textit{1.39}\spx & 2.44 & \textit{2.01}\spx & 2.53 & \textit{2.12}\spx & 2.38 & \textit{1.97}\spx & 1.90 & \textit{1.58}\spx \\
DSpark & 2.71 & \textit{2.14}\spx & 1.92 & \textit{1.53}\spx & 2.62 & \textit{2.02}\spx & 2.71 & \textit{2.23}\spx & 2.49 & \textit{2.04}\spx & 1.96 & \textit{1.60}\spx \\
\rowcolor{gray!12}\sysname & \textbf{2.94} & {\bfseries\itshape 2.50}\spx & \textbf{2.26} & {\bfseries\itshape 1.82}\spx & \textbf{2.71} & {\bfseries\itshape 2.15}\spx & \textbf{2.89} & {\bfseries\itshape 2.38}\spx & \textbf{2.69} & {\bfseries\itshape 2.21}\spx & \textbf{2.11} & {\bfseries\itshape 1.74}\spx\rule[-0.8ex]{0pt}{0pt} \\
\arrayrulecolor{black}\specialrule{\heavyrulewidth}{0pt}{0pt}\arrayrulecolor{black!42}
\end{tabular}
\end{adjustbox}\arrayrulecolor{black}
\end{table*}

As shown in Table~\ref{tab:finetuned-transfer}, \sysname remains the best drafter under both weight-only and weight-and-chat-template transfers, demonstrating the robustness of its gains to target-model shifts. Paired bootstrap 95\% CIs show that \sysname's gain over the best baseline is statistically significant in 11/12 (target transfer, metric) settings. The only exception is $\tau$ on SFT-Sci, where the CI for the gain over the best baseline, DSpark, includes zero; however, \sysname's lead over the second-best baseline, P-Eagle, is statistically significant on that setting.

\end{document}